\RequirePackage{fix-cm}
\documentclass[twocolumn]{svjour3}          

\smartqed  
\usepackage{times}
\usepackage[table,xcdraw]{xcolor} 
\usepackage{booktabs} 
\usepackage{amsmath,amsfonts}
\usepackage{algorithm}%
\usepackage{algorithmicx}%
\usepackage{array}
\usepackage[caption=false,font=normalsize,labelfont=sf,textfont=sf]{subfig}
\usepackage{textcomp}
\usepackage{stfloats}
\usepackage{url}
\usepackage{verbatim}
\usepackage{graphicx} 
\usepackage{amsmath}
\usepackage{amsfonts}
\usepackage{multirow}
\usepackage{color,xcolor}
\usepackage{bbding}
\usepackage{graphicx}
\usepackage{amsmath}
\usepackage{amssymb}
\usepackage{booktabs}
\usepackage{enumitem}
\usepackage{multirow}
\usepackage{setspace} 
\usepackage{appendix}

\usepackage{algorithm}
\usepackage{algpseudocode}

\usepackage[misc]{ifsym} 
\usepackage{newfloat}

\usepackage{listings}
\DeclareCaptionStyle{ruled}{labelfont=normalfont,labelsep=colon,strut=off} 
\floatstyle{ruled}
\newfloat{listing}{tb}{lst}{}
\floatname{listing}{Listing}

\usepackage[colorlinks,
            linkcolor=red,       
            anchorcolor=blue,  
            citecolor=blue,        
            ]{hyperref}

\usepackage{orcidlink}

\begin{document}
\begin{sloppypar}
\title{Geometric Prior Guided Feature Representation Learning \\for Long-Tailed Classification
}


\author{Yanbiao Ma, Licheng Jiao$^\textrm{\Letter}$, Fang Liu, Shuyuan Yang, Xu Liu and Puhua Chen}


\institute{Yanbiao Ma \at
              School of Artificial Intelligence \\ Xidian University, Xi'an 710071, China \\
              \email{ybmamail@stu.xidian.edu.cn}           
          \and
          Licheng Jiao \at
              School of Artificial Intelligence \\ Xidian University, Xi'an 710071, China \\
              \email{lchjiao@mail.xidian.edu.cn}
          \and
          Fang Liu \at
             School of Artificial Intelligence \\ Xidian University, Xi'an 710071, China \\
              \email{f63liu@163.com}
          \and
          Shuyuan Yang \at
              School of Artificial Intelligence \\ Xidian University, Xi'an 710071, China \\
              \email{syyang@xidian.edu.cn}
          \and
          Xu Liu \at
              School of Artificial Intelligence \\ Xidian University, Xi'an 710071, China \\
              \email{xuliu361@163.com} \\
          \and
          Puhua Chen \at
              School of Artificial Intelligence \\ Xidian University, Xi'an 710071, China \\
              \email{phchen@xidian.edu.cn} \\
        $^\textrm{\Letter}$Licheng Jiao is the corresponding authors.\\
}

\date{Accepted: IJCV 2024}

\maketitle
\begin{abstract}Real-world data are long-tailed, the lack of tail samples leads to a significant limitation in the generalization ability of the model. Although numerous approaches of class re-balancing perform well for moderate class imbalance problems, additional knowledge needs to be introduced to help the tail class recover the underlying true distribution when the observed distribution from a few tail samples does not represent its true distribution properly, thus allowing the model to learn valuable information outside the observed domain. In this work, we propose to leverage the geometric information of the feature distribution of the well-represented head class to guide the model to learn the underlying distribution of the tail class. Specifically, we first systematically define the geometry of the feature distribution and the similarity measures between the geometries, and discover four phenomena regarding the relationship between the geometries of different feature distributions. Then, based on four phenomena, feature uncertainty representation is proposed to perturb the tail features by utilizing the geometry of the head class feature distribution. It aims to make the perturbed features cover the underlying distribution of the tail class as much as possible, thus improving the model’s generalization performance in the test domain. Finally, we design a three-stage training scheme enabling feature uncertainty modeling to be successfully applied. Experiments on CIFAR-10/100-LT, ImageNet-LT, and iNaturalist2018 show that our proposed approach outperforms other similar methods on most metrics. In addition, the experimental phenomena we discovered are able to provide new perspectives and theoretical foundations for subsequent studies. The code will be available at \url{https://github.com/mayanbiao1234/Geometric-metrics-for-perceptual-manifolds}
\end{abstract}

\section{Introduction}\label{sec1}
Deep learning has made significant progress in image classification, image segmentation, and other fields benefiting from artificially annotated large-scale datasets. However, real-world data tends to follow a long-tailed distribution \cite{paper33}, and unbalanced classes introduce bias into machine learning models. Numerous approaches have been proposed to mitigate the model bias, such as class re-balancing \cite{paper6,paper12,paper50,paper53,paper27,learning}, information augmentation \cite{paper4,paper20,paper44,paper49,paper18} and network structure design \cite{paper20,paper45,paper50}. However, the above approach does not work effectively in all cases, and the generalization ability of the model will be greatly limited when the samples of the tail class do not accurately represent its true distribution. We discuss two cases of the relationship between the observed and true distributions of the tail classes \cite{paper5}.

\begin{figure}[t]
\centering
\includegraphics[width=3in]{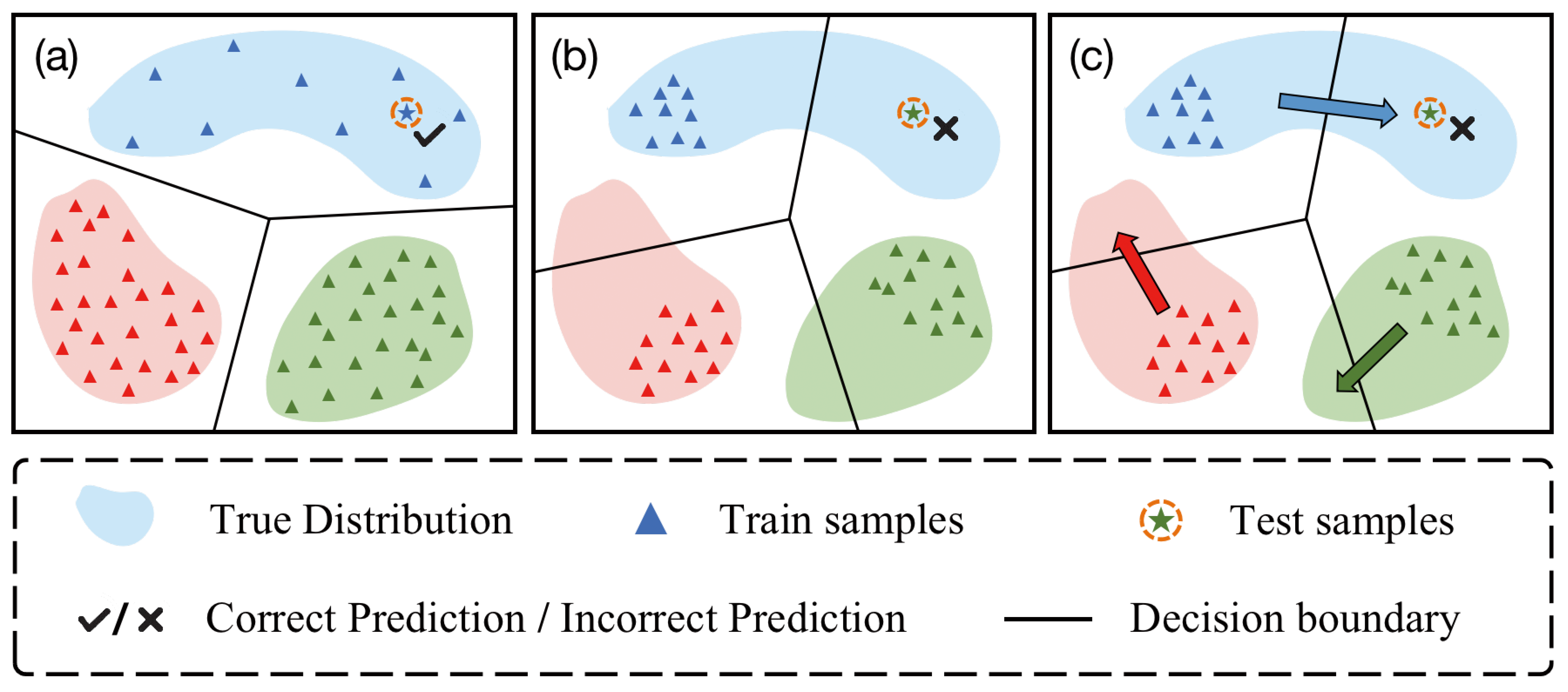}
\vskip 0.05in
\caption{(a) When the samples uniformly cover the true data distribution, the model can learn the correct decision boundaries and can correctly classify unfamiliar samples to be tested. (b) When the samples cover only a portion of the true distribution, unfamiliar samples to be tested are highly likely to be misclassified due to the error in the decision boundary. (c) The direction in which the arrow points is the best direction to expand the sample.}
\label{fig1}
\end{figure}

\begin{itemize}[]
\item Case 1: The observed samples cover the true data distribution uniformly (As shown in Figure \ref{fig1}a).
\item Case 2: The observed samples cover only a small region of the true data distribution (As shown in Figure \ref{fig2}b).
\end{itemize}

In case 1, although the sample size of the tail classes is small, these samples represent the true data distribution. The main reason for the degradation of the model performance is that at each sampling, samples from the tail class are used with a small probability to calculate loss and update parameters, resulting in inadequate learning of the tail class by the model.
Faced with this situation, existing data augmentation methods \cite{paper4,paper41}, undersampling \cite{paper36}, oversampling \cite{paper37,paper44}, and rebalancing loss \cite{paper19,paper23,paper27} can reasonably improve performance. Combining decoupled training with the above approach can improve the performance of the model even further \cite{paper1,paper45}. 
However, rebalancing strategies, including numerous meta-learning methods \cite{paper27,learning,meta-rethinking}, do not increase the information outside the observed training domain. Some meta-learning methods \cite{paper18} used for information augmentation also struggle to provide valuable information due to weak guidance from the meta-data \cite{paper42}. Additionally, CReST \cite{self-crest}, a self-training-based approach, attempts to label new samples for the tail classes, but the introduced new samples are already well-recognized by the long-tailed model, resulting in very limited new information being introduced.
In summary, when certain classes are severely underrepresented (i.e., case 2), these methods have difficulty finding the right direction for adjusting the decision boundaries, so improvements sometimes worsen \cite{paper5,paper46}. Therefore, we pursue to mine knowledge from the well-represented head classes to help recover the true distribution of the tail classes.

In case 2, if the underlying true distribution of the tail classes cannot be recovered, then the model always fails to learn the correct decision boundaries. Even if the model achieves high recognition accuracy in the training set, it still fails to have satisfactory generalization performance when faced with test samples outside the training domain. Therefore, if the direction for recovering the true distribution of tail classes, such as the direction indicated by the three arrows in Figure \ref{fig1}c, can be found, the generalization ability of the model on the tail classes will be significantly improved. It is necessary to explore additional knowledge to guide the tail class to recover the true distribution. Transfer learning for long-tail classification aims to introduce new information to facilitate the model's learning on tail classes, which can be classified into two categories \cite{paper46}: head-to-tail knowledge transfer and model pretraining. Due to the natural scarcity of tail class data, model pretraining is not effectively applicable in long-tail scenarios \cite{paper42,paper46}. Therefore, most of the current research focuses on head-to-tail knowledge transfer. However, the motivations for previous head-to-tail knowledge transfer were limited to qualitative analysis or speculation \cite{paper1,paper14,paper26,paper34}. They assumed that all background information from head classes exists in tail classes, which is an unsupported assumption lacking evidence. Moreover, these methods neglected the potential diversity of foreground information in tail classes, leading to the possibility of inadequate recovery of the true distribution of tail classes. Our experimental comparisons also demonstrate that knowledge transfer supported by evidence outperforms the aforementioned methods. In conclusion, recovering the underlying true distribution with limited samples remains a meaningful challenge.

It has been shown that the model bias is caused by the classifier and the long-tailed data does not unbalance the feature representation learning \cite{paper12,paper50}. Also considering that the dimensionality of the feature space is smaller than that of the sample space, we focus on recovering the true distribution of tail classes in the feature space. In this work, our main contributions are summarized as follows.

\begin{itemize}[]
\item We systematically define the geometry of the feature distribution and the similarity measure between the geometries (Section \ref{sec3.1} and \ref{sec3.2}). Based on this, four surprising experimental phenomena are found which can be used to guide and recover the true distribution of the tail classes (Section \ref{sec3.3}). The most important phenomenon is that similar feature distributions have similar geometries and the similarity between the geometries of the feature distributions decreases as the interclass similarity decreases. We introduce a geometric perspective to recover underrepresented class distributions, providing a theoretical and experimental basis for subsequent studies of class imbalance.
\item Based on four experimental phenomena, we propose to model the uncertainty representation of the tail features with geometric information from the feature distribution of the head class (Section \ref{sec4.1}). Specifically, instead of treating samples in the feature space as deterministic points, we perturb them to make the model learn information outside the observed domain by taking into account the geometry of the class to which the samples belong. Our proposed feature uncertainty modeling can effectively alleviate the model bias introduced by underrepresented classes and can be easily integrated into existing networks.
\item We propose a three-stage training scheme to apply feature uncertainty representation (Section \ref{sec4.2}). The results of the ablation experiments show that compared to decoupled training, the three-stage training scheme improves the tail class performance while reducing the degradation of the head class performance, resulting in more overall performance improvement of the model.
\item Experiments on large-scale long-tailed datasets (Section \ref{sec5}) show that our proposed method significantly improves the performance of tail classes and exhibits state-of-the-art results compared to other similar methods.
\end{itemize}

\section{Related Work}\label{sec2}
\subsection{Class Rebalancing}\label{sec2.1}

The extreme imbalance in the number of samples in the long-tail data prevents the classification model from learning the distribution of the tail classes adequately, which leads to poor performance of the model on the tail classes. Therefore, methods to rebalance the number of samples and the losses incurred per class (i.e., resampling and reweighting) are proposed. Resampling methods are divided into oversampling and undersampling \cite{paper3,paper8,paper9,paper12,paper35,paper47}. The idea of oversampling is to randomly sample the tail classes to equalize the number of samples and thus optimize the classification boundaries. The undersampling methods balance the number of samples by randomly removing samples from the head classes. For example, \cite{paper36} finds that training with a balanced subset of a long-tailed dataset is instead better than using the full dataset.
In addition, \cite{paper12,paper50} fine-tune the classifier via a resampling strategy in the second phase of decoupled training. \cite{paper37} continuously adjusts the distribution of resampled samples and the weights of the two-loss terms during training to make the model perform better. \cite{paper44} employs the model classification loss from an additional balanced validation set to adjust the sampling rate of different classes.

The purpose of reweighting loss is intuitive, and it is proposed to balance the losses incurred by all classes, usually by applying a larger penalty to the tail classes on the objective function (or loss function) \cite{paper7,paper10,paper30,paper31,paper38,paper48,paper52,paper53}. \cite{paper27} proposes to adjust the loss with the label frequencies to alleviate class bias. \cite{paper19} not only assigns weights to the loss of each class but also assigns higher weights to hard samples. Recent studies have shown that the effect of reweighting losses by the inverse of the number of samples is modest \cite{paper23,paper25}. Some methods that produce more ``smooth" weights for reweighting perform better \cite{paper53}, such as taking the square root of the number of samples as the weight \cite{paper23}. \cite{paper6} attributes the better performance of this smoother method to the existence of marginal effects. In addition, \cite{paper1} proposes to learn the classifier with class-balanced loss by adjusting the weight decay and MaxNorm in the second stage. DSB \cite{paper53} and DCR \cite{paper52}, for the first time, examined the factors influencing model bias from a geometric perspective and proposed a rebalancing approach.

Although class rebalancing methods are simple to implement, their limitations have been increasingly recognized in recent research \cite{paper46,paper54}. Class rebalancing methods merely increase the weight of the tail class loss without introducing additional knowledge to assist the tail classes, which often leads to overfitting of the tail classes and significantly compromises the model's generalization performance \cite{paper54}. Another limitation is that class rebalancing methods often improve tail class performance at the expense of sacrificing head class performance, making it challenging to handle data scarcity issues \cite{paper46, paper42}. As a result, more and more research is focusing on information augmentation.

\subsection{Stage-wise training}\label{sec2.2}

Decoupling \cite{paper12} first proposes to decouple the learning process on long-tail data into feature learning and classifier learning, and it finds that re-learning the balanced classifier can significantly improve the model performance. Further, BBN \cite{paper50} combines the two-stage learning into a two-branch model. The two branches of the model share parameters, with one branch learning using the original data and the other learning using the resampled data. \cite{paper5} decomposes the features into class-generic features and class-specific features, and it expands the tail class data by combining class-generic features of the head class with class-specific features of the tail class. \cite{paper49} finds that augmenting data with Mixup in the first stage benefits feature learning and does negligible damage to classifiers trained using decoupling. \cite{paper45} also observes that long-tailed data does not affect feature learning, and it proposes an adaptive calibration function for improving the cross-entropy loss. \cite{paper11} considers the effect of noisy samples on the tail class and adaptively assigns weights to the tail class samples by meta-learning in the second stage. 


The two-stage training pushes the decision boundary away from the augmented tail class distribution, thereby improving the performance of the tail classes. However, this may lead to excessive bias in the decision boundary and affect the head classes \cite{paper43}. Therefore, we propose a three-stage training strategy. The first two stages are indistinguishable from decoupled training, while in the third stage, we fix the classifier parameters and fine-tune the feature extractor to adapt it to the improved classification boundary.

\subsection{Head-to-tail knowledge transfer}\label{sec2.3}

\textbf{Head-to-tail knowledge transfer is more relevant to our work than other methods.} \cite{paper43} and \cite{paper21} were first proposed in the face recognition field to transfer variance between classes to augment classes with fewer samples. \cite{paper43,paper55} assumes that the feature distributions of each class are multivariate Gaussian, and the feature distributions of the common and under-represented classes have the same variance, the variance of the head class is used to estimate the distribution of the tail class. \cite{paper21} assumes that the intra-class angle distribution follows a Gaussian distribution, transfers the intra-class angle distribution of features to the tail class, and constructs a ``feature cloud" for each feature to extend the distribution of the tail class. Both \cite{paper43} and \cite{paper21} assume that certain statistical properties of head and tail classes are the same (i.e., variance and intra-class angular distribution). However, these assumptions lack supporting evidence, which limits their performance in long-tail recognition.

Similar to the adversarial attack, \cite{paper14} proposes to transform some of the head samples into tail samples through perturbation-based optimization to achieve tail class augmentation. However, this method exploits the vulnerabilities of deep neural networks to generate samples that mislead the model, and these samples do not exist as tail-class samples in reality. As a result, it cannot effectively help the tail classes recover their underlying distribution. \cite{paper5} decomposes the features of each class into class-generic features and class-specific features. During training, the class-specific features of the tail class are fused with the generic features of the head class to generate new features to expand the tail class. This idea is similar to data augmentation in image space, such as Cutmix. \cite{paper34} dynamically estimates a set of centers for each class, and then calculates the displacement between the head class feature and the corresponding nearest intra-class center. This displacement is used to combine with the tail class centers to generate new features, thereby increasing the feature diversity of the tail class. \cite{paper20} proposes to transfer the geometric information of the feature distribution boundaries of the head class to the tail class by enhancing the weights of the tail class classifier. The recently proposed CMO \cite{paper26} considers that the image of the head class has a rich background, so the image of the tail class can be pasted directly onto the background image of the head class to increase the richness of the tail class. Overall, the above-mentioned research assumes that all background information of head classes exists in tail classes, which is an unsupported assumption. It also overlooks the potential diversity of tail class foreground information, which may lead to ineffective recovery of the true distribution of tail classes.

Distinguishing from the above studies, we pioneered a geometric perspective of head-tail knowledge transfer. We systematically define the geometry of the distribution and its similarity measure and find direct evidence that the geometry of the head class distribution can help the tail class.

\section{Motivation}\label{sec3}

We first define a measure of the geometry of the feature distribution, and then propose a similarity measure between the geometries. Finally, across several benchmark data sets, we discovered four experimental phenomena regarding the relationship between geometric information of feature distributions. Inspired by the experimental phenomenon, we propose to utilize the feature distribution of the head class to help the tail class to recover the underlying distribution.

\subsection{The Geometry of Data Distribution}\label{sec3.1}
In the $P$-dimensional space, given data $X=[x_1,x_2,\dots,x_n]\in \mathbb{R}^{P\times n}$ that belongs to the same class, the sample covariance matrix of $X$ can be estimated as
\begin{equation}
\begin{split}
\Sigma_X=\mathbb{E}[\frac{1}{n}\sum_{i=1}^{n}x_ix_i^T]=\frac{1}{n}XX^T\in \mathbb{R}^{P\times P}.
\nonumber
\end{split}
\end{equation}

If $\Sigma_X=I_P$ and $I_P$ denotes a unit matrix of order $P$, the distribution of $X$ is said to be isotropic, while the opposite is said to be anisotropic. In practice, the data distribution is usually anisotropic.
Considering the two-dimensional case, we can find two vectors $\xi_1$ and $\xi_2$, where $\xi_1$ points to the direction with the largest sample variance, and $\xi_2$ points to the direction with the largest variance among the directions orthogonal to $\xi_1$. $\xi_1$ and $\xi_2$ can be used to anchor the geometry of the two-dimensional distribution. In the high-dimensional case, since $\Sigma_X$ is a real symmetric matrix, any two of its eigenvectors are orthogonal to each other, and $\xi_i$ points to the direction with the $i$-th largest variance. Analogously to the two-dimensional case, we can use all the eigenvectors of $\Sigma_X$ to anchor the geometry of the distribution.

\begin{definition}[\textbf{The geometry of data distribution}]\label{def1}
Given a $P$-dimensional sample set $X$ and the corresponding covariance matrix $\Sigma_X$. The eigendecomposition of $\Sigma_X$ yields $P$ eigenvalues $\{ \lambda_1,\lambda_2,\dots,\lambda_P$ and the corresponding $P$-dimensional eigenvectors $[\xi_1,\xi_2,\dots,\xi_P]\in \mathbb{R}^{P\times P}$. All eigenvectors of $\Sigma_X$ are considered as bones to anchor the geometry of the distribution of $X$, denoted as
\begin{equation}
\begin{split}
GD_X(\xi_1,\xi_2,\dots,\xi_P),
\nonumber
\end{split}
\end{equation}
where $\lambda_1\ge \lambda_2\ge\dots\ge\lambda_P\ge0$, $\Vert \xi_i \Vert_2=1$, $i=1,2,\dots,P$.
\end{definition}

\subsection{Similarity Measure of Geometry}\label{sec3.2}
In the $P$-dimensional space, given two types of data $X_1=[x_1,\dots,x_n]\in \mathbb{R}^{P\times n}$ and $X_2=[x_1,\dots,x_n]\in \mathbb{R}^{P\times m}$,  their sample covariance matrices are estimated as $\Sigma_{X_1}=\frac{1}{n}X_1X_1^T\in \mathbb{R}^{P\times P}$ and $\Sigma_{X_2}=\frac{1}{m}X_2X_2^T\in \mathbb{R}^{P\times P}$, respectively. Performing the eigendecomposition on $\Sigma_{X_1}$ and $\Sigma_{X_2}$, the geometry of the distributions $X_1$ and $X_2$ are denoted as $GD_{X_1}(\xi_{X_1}^1,\dots,\xi_{X_1}^P)$ and $GD_{X_2}(\xi_{X_2}^1,\dots,\xi_{X_2}^P)$, respectively, where $\xi_{X_1}^i$ and $\xi_{X_2}^j (i,j=1,2,\dots,P)$ are the eigenvectors of $\Sigma_{X_1}$ and $\Sigma_{X_2}$, respectively.

\begin{definition}[\textbf{Similarity metric between geometry}]\label{def2}
Given the geometry of two distributions $GD_{X_1}(\xi_{X_1}^1,\dots,\xi_{X_1}^P)$ and $GD_{X_2}(\xi_{X_2}^1,\dots,\xi_{X_2}^P)$, their similarity is defined as 
\begin{equation}
\begin{split}
S(GD_{X_1},GD_{X_2}) = \sum_{i=1}^{P} \left |  \langle \xi_{X_1}^i,\xi_{X_2}^i \rangle \right | = \sum_{i=1}^{P} \left |{\xi_{X_1}^i}^T\xi_{X_2}^i \right | . 
\nonumber
\end{split}
\end{equation}
\end{definition}

The larger $S(GD_{X_1},GD_{X_2})$, the more similar the geometry of the distributions $X_1$ and $X_2$. The upper and lower bounds of $S(GD_{X_1},GD_{X_2})$ are
\begin{equation}
\begin{split}
0\le S(GD_{X_1},GD_{X_2}) \ge P.
\nonumber
\end{split}
\end{equation}
 
When any  pair of eigenvectors $\xi_{X_1}^i$ and $\xi_{X_2}^i$ are co-linear, $S(GD_{X_1},GD_{X_2})$ reaches the upper bound $P$. When any pair of eigenvectors $\xi_{X_1}^i$ and $\xi_{X_2}^i$ are orthogonal, $S(GD_{X_1},GD_{X_2})$ takes the lower bound value $0$. Taking the two-dimensional distribution as an example, since
\begin{small}
\begin{equation}
\begin{split}
0\le {\phi_{R1}}^T\phi_{B1}+{\phi_{R2}}^T\phi_{B2}\le {\xi_{R1}}^T{\xi_{B1}}+{\xi_{R2}}^T{\xi_{B2}}\le 2,
\nonumber
\end{split}
\end{equation}
\end{small}
it is clear that the geometry of the two distributions in Figure \ref{fig6} is more similar compared to the two distributions shown in Figure \ref{fig7}. The details are described in Appendix \ref{secA}.

\subsection{Four Discoveries about the Geometry of the Feature Distribution}\label{sec3.3}
First define the class similarity measure. Then introduce the four phenomena we found and their experimental setup.

\begin{definition}\label{def3}
Given a sample set $D_c=\{ \dots,(x_i,y_c),\dots \}$ of class $c$, the average prediction score $\frac{1}{\mid D_c \mid}  {\textstyle \sum_{i}}p(y_c\mid x_i,\theta)$ of all samples belonging to class $c$ is calculated using a deep neural network with trained parameters $\theta$, where $\mid \!\!D_c\!\! \mid$ denotes the sample number of class $c$. Define the class
\begin{equation}
\begin{split}
h:=argmax_{k\neq c}(\frac{1}{\mid D_c \mid}  {\textstyle \sum_{i}}p(y_c\mid x_i,\theta) )_k
\nonumber
\end{split}
\end{equation}
that is most similar to class $c$, i.e., the class with the largest logit other than class $c$. Further, the similarity ranking can be done based on logit.
\end{definition}

We investigated the relationship between class similarity and the geometry similarity of class distributions on two benchmark datasets: Fashion MNIST \cite{paper39} and CIFAR-10 \cite{paper15}. ResNet-18 \cite{paper40} was adopted as the backbone network and various training schemes were applied to make the performance of ResNet-18 on the two datasets comparable to the state-of-the-art results (See Appendix \ref{secB} for details). First, the similarity between all classes on the two datasets is calculated and ranked. Then, we extracted $64$-dimensional features of all samples from both datasets using ResNet18 and calculated the geometry of all class feature distributions. Based on this, we summarize further experiments and findings as follows.

\subsubsection{Phenomenon 1}\label{sec3.3.1}
\begin{figure}[t]
\centering
\includegraphics[width=3in]{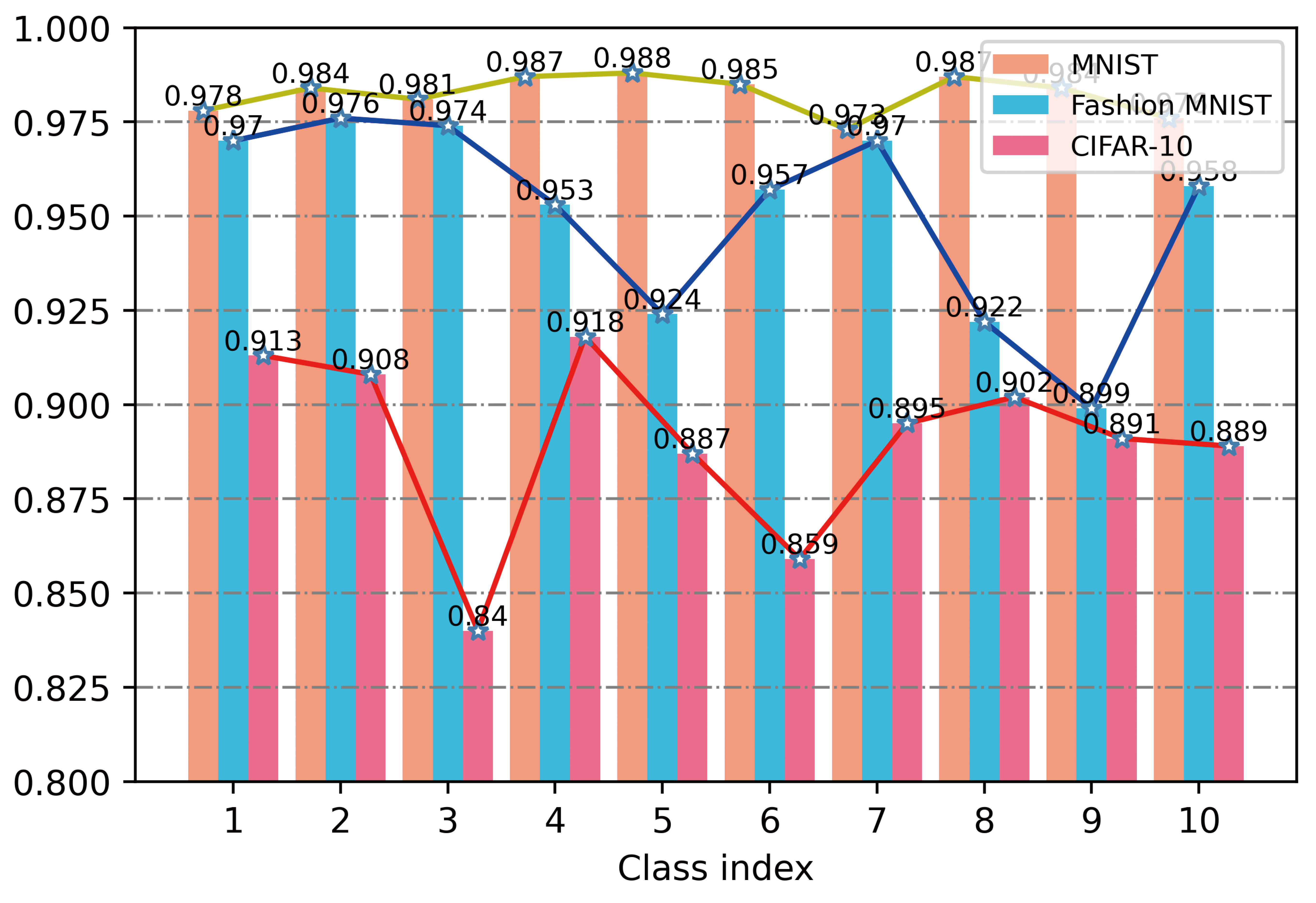}
\vskip 0.05in
\caption{The ratio of the sum of the top five eigenvalues to the sum of all eigenvalues after eigendecomposition for the feature embeddings of all classes in the three datasets. The horizontal coordinates are the indexes of the classes, and the specific class names are in Appendix \ref{secD}.}
\label{fig2}
\vskip -0.05in
\end{figure}

As shown in Figure \ref{fig2}, features were extracted using trained ResNet-18 on MNIST \cite{paper16}, Fashion MNIST and CIFAR-10. We find the sum of the eigenvalues corresponding to the first five eigenvectors that are used to represent the geometry of the distribution can reach more than $80\%$ of the sum of all eigenvalues, which means that most of the information of the data distribution can be recovered along the first five eigenvectors.

\subsubsection{Phenomenon 2}\label{sec3.3.2}
\begin{figure}[t]
	\centering
       \begin{minipage}{1\linewidth}
		\centering
		\includegraphics[width=1\linewidth]{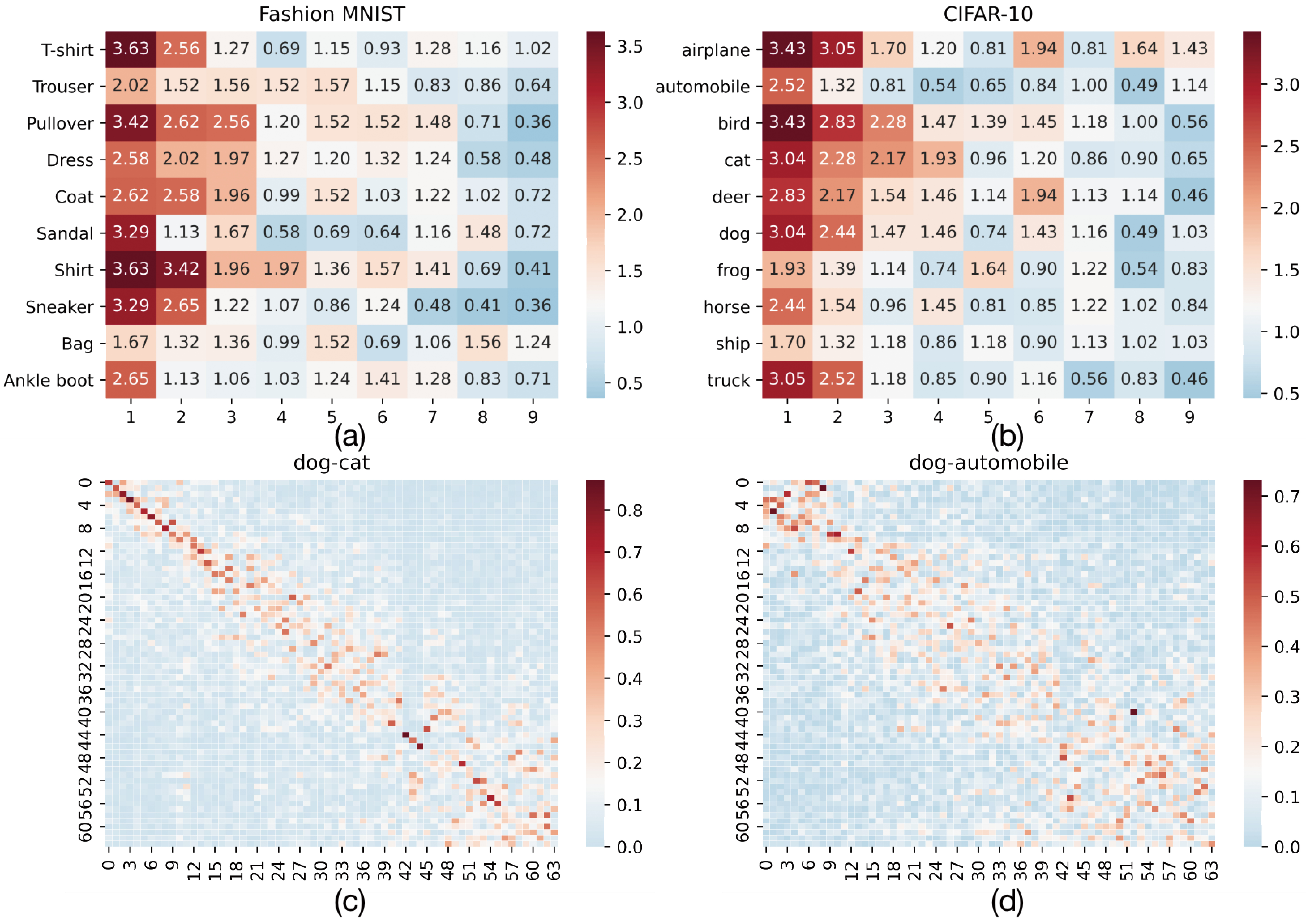}
	\end{minipage}

	\begin{minipage}{0.493\linewidth}
		\centering
		\includegraphics[width=1\linewidth]{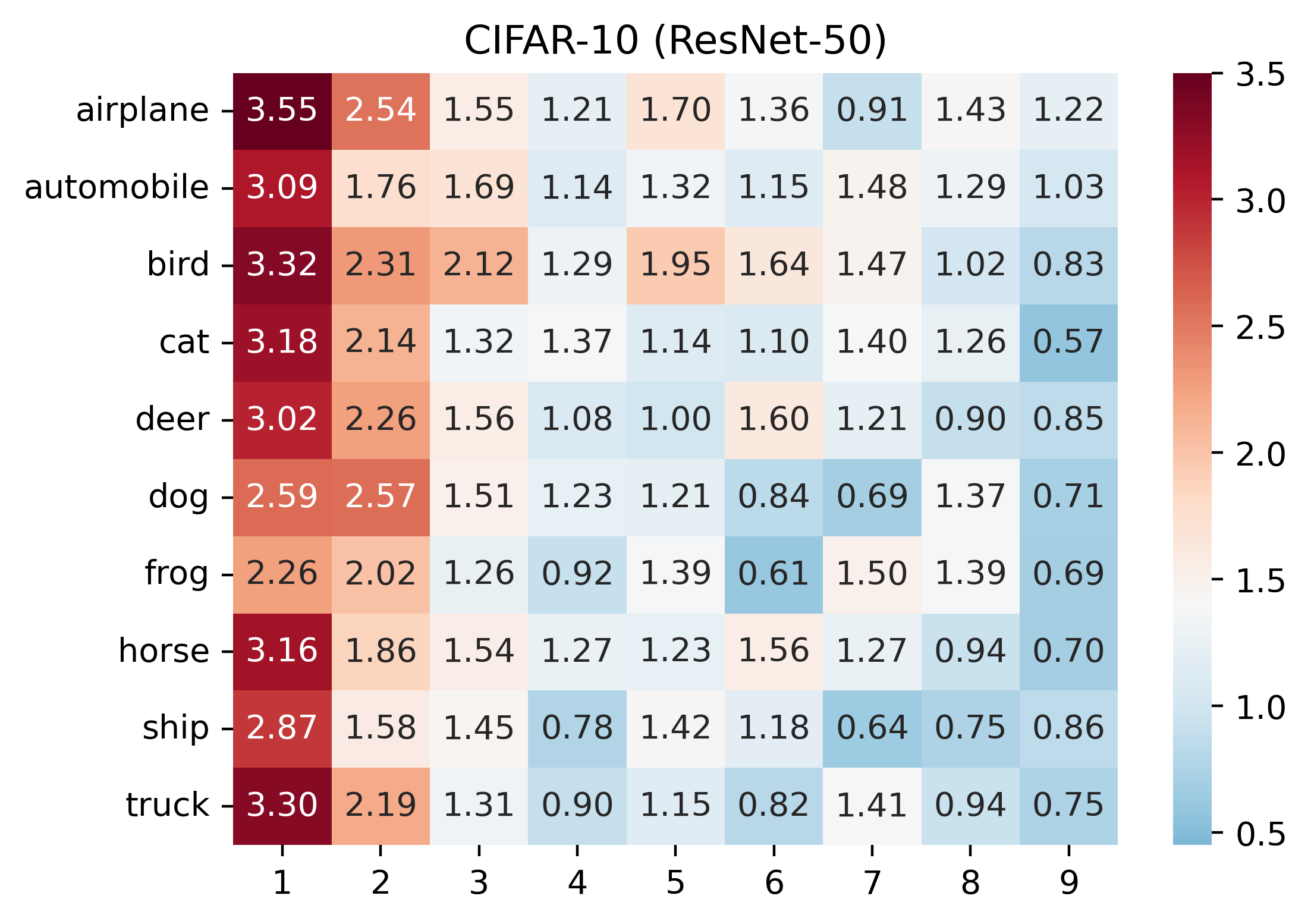}
	\end{minipage}
	\begin{minipage}{0.493\linewidth}
		\centering
		\includegraphics[width=1\linewidth]{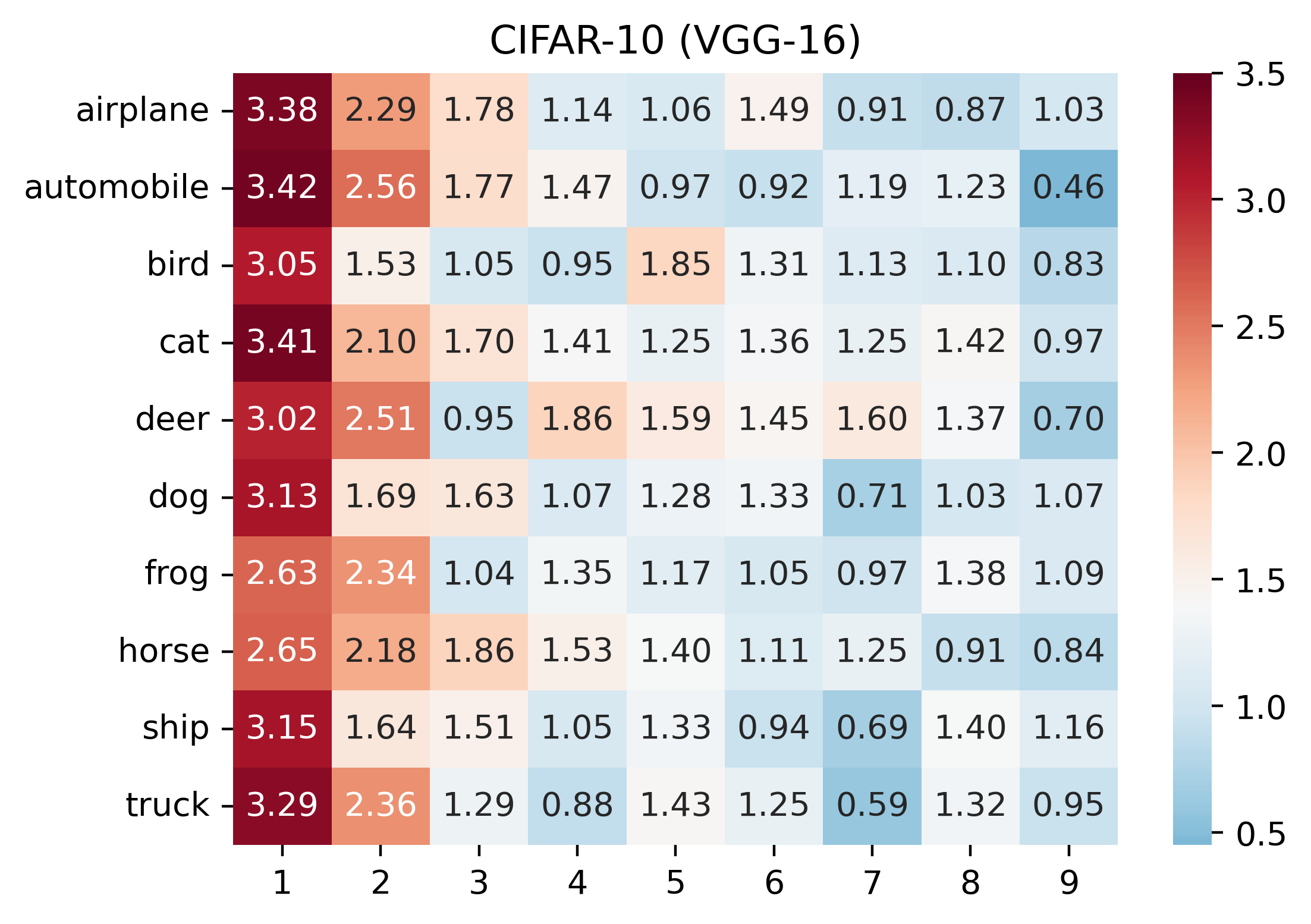}
	\end{minipage}
\caption{(a) The horizontal coordinates are the indexes of the classes, and 1 to 9 indicate the classes that are most similar to the class represented by the vertical coordinates to the least similar, respectively. Each element represents the similarity of the geometry between classes. See Appendix \ref{secD} for detailed class names. (b) Same as (a). (c) The inner product between all eigenvectors of \emph{dog} and all eigenvectors of \emph{cat} in CIFAR-10. The sum of the first five diagonal elements of $M1$ is equal to the value of the element in the first column of the first row in (b). (d) The inner product between all eigenvectors of \emph{dog} and \emph{automobile} in CIFAR-10.
The third row represents the results of the experiments using ResNet-50 and VGG-16 as backbone networks. The axes as well as the meanings of the values are consistent with (a) and (b).}
\label{fig3}
\end{figure}

Based on the above observations, we set $P$ in $S(GD_{X_1}, GD_{X_2})$ to $5$ and calculate the similarity between geometry of all class feature distributions in Fashion MNIST and CIFAR-10 and plot them in Figure \ref{fig3}a and Figure \ref{fig3}b. We find that \textbf{if two classes have high similarity, then the geometry of their feature distributions also exhibit high similarity}. And as the similarity between classes decreases, the similarity between the geometry of the class feature distributions shows a decreasing trend. Take \!\emph{dog}\! in CIFAR-10 as an example, its most and least similar classes are \emph{cat} and \emph{automobile}, respectively, and the geometry of the three classes are represented by $GD_{airplane}(\xi_1,\dots,\xi_{64})$, $GD_{bird}(\eta_1,\dots,\eta_{64})$ and $GD_{horse}(\zeta_1,\dots,\zeta_{64})$. Calculate the matrices \!$M1$\! and \!$M2$\! and plot them in Figure \ref{fig3}c and Figure \ref{fig3}d, where $M1_{i,j} \!\!= \!\! \langle \xi_i, \eta_j \rangle$ and $M2_{i,j} \!\!= \!\! \langle \xi_i, \zeta_j \rangle (i,j=1,\dots,64)$. It can be observed that \!$M1$\! is closer to a diagonal matrix compared to \!$M2$, which corresponds to a more similar geometry of \!\emph{dog}\! and \!\emph{cat}. 
Furthermore, we validate our findings using ResNet-50 and VGG-16 as backbone networks on CIFAR-10. The experimental results are shown in the third row of Figure \ref{fig3}, and it can be observed that the phenomenon we discovered still holds under different backbone networks.

To prove that the above phenomenon does not occur by chance, we give the probability that the experimental results in Figure \ref{fig3}c occur randomly. Given two random vectors in a $P$-dimensional space, let their inner product be $\delta \in [0,1]$. The probability density function of $\delta$ is represented as
\begin{equation}
\label{equa1}
\begin{split}
f_P(\delta)=\frac{\Gamma (\frac{P}{2})}{\Gamma (\frac{P-1}{2})\sqrt{\pi} }(1-\delta^2)^{\frac{P-3}{2}}.
\end{split}
\end{equation}

\begin{figure}[t]
\centering
\includegraphics[width=3.1in]{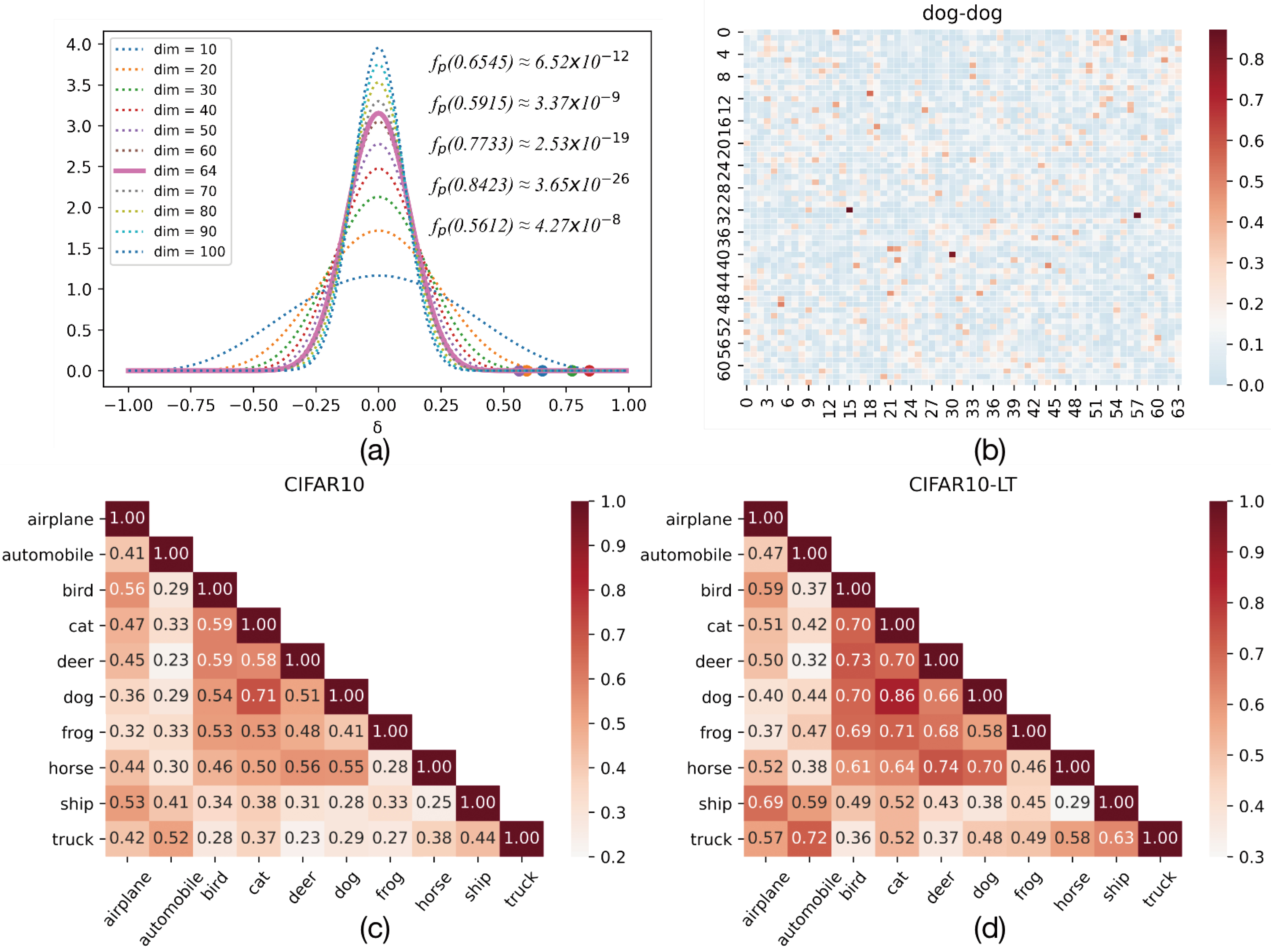}
\vskip 0.05in
\caption{(a) The function curve of Equation (\ref{equa1}). It can be observed that as the dimensionality increases, any two random vectors tend to be orthogonal to each other. (b) When two different models are used to extract features of \emph{dog} separately, the geometry of the two feature distributions is not similar. (c) Cosine similarity between feature centers of classes on CIFAR-10. (d) Cosine similarity between feature centers of classes on CIFAR-10-LT.}
\label{fig4}
\vskip -0.05in
\end{figure}

The detailed derivation and proof process of the above equation is shown in the Appendix \ref{secC}. Setting $P$ in $f_P(\delta)$ to $64$, when $\delta$ is taken as the first five diagonal elements of $M1$ respectively, the calculation result of $f_P(\cdot)$ is shown in Figure \ref{fig4}a. Considering only the first five diagonal elements of $M1$, the probability of the situation shown in Figure \ref{fig3}c occurring is almost $0$. Not only that, we observed numerous such phenomena (see Appendix \ref{secD}), thus implying that the phenomena we found could hardly have occurred by chance.

\subsubsection{Phenomenon 3}\label{sec3.3.3}

The phenomenon that features distributions of a similar class has similar geometry only occurs when all features are extracted using the same model. Figure \ref{fig4}b shows that there is a low similarity between the geometry of \emph{dog} computed by two different ResNet18 trained with random initialization. More examples in Appendix \ref{secE}.

\subsubsection{Phenomenon 4}\label{sec3.3.4}

We conducted further experiments on CIFAR-10 as well as its long-tailed version CIFAR-10-LT. In CIFAR-10-LT, \emph{airplane}, \emph{automobile}, \emph{bird}, and \emph{cat} are considered head classes and the remaining classes are tail classes. As shown in Figure \ref{fig4}d, we confirmed that the most similar class to the tail class usually belongs to the head class \cite{paper5} and found that if a tail class and a head class show high similarity in CIFAR-10-LT, they also show high similarity on CIFAR-10.

\subsubsection{Summary and inspiration}\label{sec3.3.5}

Combining the above four phenomena, we propose the following idea: the most similar head class is selected for each tail class in the training process, and the geometry of the head class feature distribution is taken as a priori knowledge to guide and recover the underlying true distribution of the tail class.

\section{Methodology}\label{sec4}
We first introduce how to leverage the geometric information of the head class feature distribution to model the uncertainty representation of the tail class features, allowing the model to learn the underlying true distribution of the tail class. Then a three-stage training scheme is proposed to apply the feature uncertainty representation.

\subsection{Feature Uncertainty Representation}\label{sec4.1}

Given a tail class $t$, the head class that is most similar to class $t$ is assumed to be $h$. The $P$-dimensional feature embedding belonging to tail class $t$ is $z_t=[z_t^1,\dots,z_t^{N_t}]^T\in \mathbb{R}^{P\times N_t}$ and the feature embedding belonging to head class $h$ is $z_h=[z_h^1,\dots,z_h^{N_h}]^T\in \mathbb{R}^{P\times N_h}$, where $N_t$ and $N_h$ denote the sample numbers of class $t$ and class $h$, respectively. The $i$-th feature embedding of $z_t$ is denoted by $z_t^i$. For the model to learn the underlying distribution of the tail class $t$, we want to utilize the existing feature embeddings to generate feature embeddings that can cover the underlying distribution of the tail class $t$. We therefore propose to model the uncertainty representation of $z_t^i$ with the geometry of the feature distribution of class $h$, i.e., $z_t^i$ is no longer considered a deterministic point.

The sample covariance matrix of class $h$ is estimated as $\Sigma_h=\frac{1}{N_h}z_hz_h^T\in \mathbb{R}^{P\times P}$. The eigenvalues of the matrix $\Sigma_h$ are denoted as $[\lambda_h^1,\dots,\lambda_h^P]\in \mathbb{R}^P$, where $\lambda_h^1\ge \cdots \ge \lambda_h^P$. The eigenvector $[\xi_h^1,\dots,\xi_h^P]\in \mathbb{R}^{P\times P}$, which corresponds one-to-one with the eigenvalues, anchors the geometry of the class $h$ feature distribution, where $\Vert \xi_h^i \Vert_2=1, i=1,\dots,P$. Since the distributions of similar class have similar geometry, we propose to represent the uncertainty of $z_t^i$ by centering a single feature embedding $z_t^i$ of the tail class $t$ and performing a random translation to $z_t^i$ along a random linear combination of $\xi_h^1,\dots,\xi_h^P$.
Considering that the ``scope" of the distribution is larger in the direction with larger eigenvalues \cite{paper51}, an additional weight $\lambda_h^i$ is assigned to $\xi_h^i (i=1,\dots,P)$ when the eigenvectors are randomly combined, which means that $z_t^i$ is translated farther with higher probability in the direction with larger eigenvalues. In summary, the final form of the proposed method can be represented as
\begin{equation}
\begin{split}
FUR(z_t^i)&= \! \! \! \! \! \! \! \!\! \overbrace{z_t^i+\sum_{j=1}^{P}\epsilon_j\lambda_h^j\xi_h^j\in \mathbb{R}^P }^{Uncertainty \hspace{0.8mm} representation\hspace{0.8mm} of\hspace{0.8mm} z_t^i} \\
\epsilon_j&\sim N(0,1), j=1,\dots,P.
\end{split}
\end{equation}

$\epsilon_1,\dots,\epsilon_P$ all follow the standard Gaussian distribution and are independent of each other, and sampling them randomly multiple times can produce new feature embeddings with different translation directions and distances. In particular, when $\epsilon_1=1,\epsilon_2,\dots,\epsilon_P=0$, $z_t^i$ is translated along $\xi_h^1$ by a distance $\lambda_h^1$. And so on, the maximum translation distances of $z_t^i$ in the direction represented by each feature vector individually are $\lambda_h^1,\dots,\lambda_h^P$, respectively.

\begin{algorithm*}[t]
  \caption{Feature Uncertainty Representation}
  \label{alg1}
  \begin{algorithmic}[1]
    \Require A long-tailed dataset $D$ containing $S$ samples. A CNN network $M=\{f(x,\theta_1),g(z,\theta_2) \}$, where $\theta_1$ and $\theta_2$ denote the parameters of the feature sub-network and classifier, respectively, and $x$ and $z$ denote the input and feature embedding of the model, respectively.
\For {$epoch=1$ to $m1$}
\State Training model $M$ on dataset $D$ without using any class rebalancing method.
\EndFor
\State Using $M$, the head classes that are most similar to each tail class are calculated, and then the sample covariance matrix of these head classes is calculated.

\For {$epoch=m1$ to $m2$}
	\State Freeze the parameters $\theta_1$ of the feature sub-network.
	\For {$iteration=0$ to $\frac{S}{batch \hspace{0.8mm} size}$}
		\State {A mini-batch $\{(x_i,y_i)\}_{i=1}^{batch\hspace{0.8mm}  size}$ is sampled from $D$, where the sample numbers from the tail class are $N_T$ and the sample numbers from the head class are $N_T(1+N_A)$.}
		\State Compute the feature embedding: $z_i=f(x_i,\theta_1), i=1,\dots,(2N_T+N_TN_A)$.
		\State Uncertainty representation of all features from tail classes: $FUR(z_t^i)=z_t^i+ {\textstyle \sum_{j=1}^{P}}\epsilon_j \lambda_h^j \xi_h^j \in \mathbb{R}^P$, $t\in tail \hspace{0.8mm}  class,i\in int[0,N_t]$. $h$ denotes the head class most similar to $t$.
		\State $\epsilon_j \sim N(0,1), j=1,\dots,P$. $N_A$ augmented features are generated for the true features of each tail class by randomly sampling $N_A$ times of $\epsilon_j(1,\dots,P)$.
		\State A mini-batch with a balanced distribution containing $2N_T(1+N_A)$ samples is obtained.
		\State Compute the cross-entropy loss $L(g(z_i,\theta_2),y_i)$ and update the parameters of the classifier: $\theta_2=\theta_2-\alpha \nabla_{\theta_2}L(g(z_i,\theta_2),y_i)$.
\EndFor
\EndFor

\For {$epoch=m2$ to $m3$}
	\State Freeze the parameter $\theta_2$ of $g(z,\theta_2)$.
	\State Fine-tuning the parameters of the feature sub-network using the long-tailed dataset $D$.
\EndFor
  \end{algorithmic}
\end{algorithm*}

Our proposed method can be integrated as a flexible module after the feature sub-network. It generates augmented samples of tail classes in the feature space to cover the underlying distribution, giving the model better generalization ability on long-tailed data. \textbf{Note that this module is only applied during model training and can be discarded during testing without affecting the inference speed}.

\subsection{Training Scheme}\label{sec4.2}
We propose a three-stage training scheme to apply feature uncertainty representation so that the model learns information outside the observed domain. Decoupled training is adopted for the first two phases. In Phase 1, the long-tailed dataset is used to learn the feature sub-network and classifier. In Phase 2, the uncertainty representation of the tail feature is applied to generate new samples for reshaping the decision boundaries. Unlike decoupled training, we additionally add Phase 3 to fine-tune the feature sub-network to adapt it to the new decision boundaries.

\begin{itemize}[]
\item \textbf{Phase-1: Initialization training.} Represent an end-to-end deep neural network as a combination of a feature sub-network and a classifier: $M=\{f(x,\theta_1),g(z,\theta_2)\}$, where $\theta_1$ and $\theta_2$ are the parameters of the network. We utilize all images from the dataset to learn the feature sub-network $f(x,\theta_1)$ as well as the classifier $g(z,\theta_2)$. After training is completed, the head classes that are most similar to each tail class are selected based on the average prediction score of the model (see Definition \ref{def3}), and the geometry of the feature distribution of these head classes is represented by the eigenvectors of the covariance matrix, which will be applied to guide the recovery of the tail class distribution.
\item \textbf{Phase-2: Reshaping decision boundaries.} Freeze the parameters of $f(x,\theta_1)$ and employ feature uncertainty representation in feature space for the tail class to fine-tune the classifier to improve the performance of the tail class. Specifically, in each iteration, we randomly sample $N_T$ images from the tail class, and then generate $N_A$ augmented samples for each true sample by feature uncertainty representation. Meanwhile, to balance the sample distribution, we directly sample $N_T(1+N_A)$ samples randomly from the head class. The tail class samples and the head class samples together form a mini-batch containing $2N_T(1+N_A)$ samples for fine-tuning the classifier. The $N_A$ and $N_T$ settings are related to the batch size, and they are described in detail in Section \ref{sec5.2}.
\item \textbf{Phase-3: Fine-tuned feature sub-network.} Fine-tuning the decision boundary can improve the performance of the tail class while compromising the performance of the head class \cite{paper43}. This is because the feature sub-network is not well adapted to the new decision boundary. Therefore, we propose to freeze the parameters of $g(z,\theta_2)$ at Phase-3 and fine-tune $f(x,\theta_1)$ with the original long-tailed data.
\end{itemize}
The above three-phase training process is summarized in Algorithm \ref{alg1}.

\section{Experiments}\label{sec5}

\subsection{Datasets and Evaluation Metrics}\label{sec5.1}
We evaluate the effectiveness and generalizability of our approach at CIFAR-10/100-LT \cite{paper2,paper15}, ImageNet-LT \cite{paper22}, iNaturalist 2018 \cite{paper32} and \textcolor{blue}{OIA-ODIR} \cite{benchmark}. For a fair comparison, the training and test images of all datasets are officially split \cite{paper42,paper46,paper53}, and the Top-1 accuracy on the test set is utilized as a performance metric.

\begin{itemize}[]
\item Both CIFAR-10 and CIFAR-100 \cite{paper15} contain $60, 000$ images, of which $50, 000$ are used for training and $10, 000$ for validation, and they contain $10$ and $100$ classes, respectively. For a fair comparison, we use the long-tailed version of the CIFAR dataset. The imbalance factor (IF) is defined as the value of the number of the most frequent class training samples divided by the number of the least frequent class training samples. The imbalance factors we employ in our experiments are $10$, $50$, $100$, and $200$.
\item \textbf{ImageNet-LT} is an artificially produced unbalanced dataset utilizing its balanced version (ImageNet-LT-2012 \cite{paper28}). It with an imbalance factor of $256$, contains $1000$ classes totaling $115.8k$ images, with a maximum of $1280$ images and a minimum of $5$ images per class.
\item The \textbf{iNaturalist 2018} dataset is a large-scale real-world dataset that exhibits a long tail. It contains $437, 513$ training samples from $8, 142$ classes with an imbalance factor of $500$ and three validation samples per class.
The OIA-ODIR dataset was made public in $2019$, and it contains a total of $10,000$ fundus images in $8$ classes. See Section \ref{sec5.7} for a more detailed description.
\end{itemize}

\begin{table*}[t]
\caption{Comparison on CIFAR-10-LT and CIFAR-100-LT. The accuracy (\%) of Top-1 is reported. The best and second-best results are shown in \underline{\textbf{underlined bold}} and \textbf{bold}, respectively. FUR-Decoupled indicates a FUR with decoupled training, and FUR Default indicates a FUR with three-stage training scheme.}
\label{table1}
\centering  
\renewcommand\arraystretch{1}
\setlength{\tabcolsep}{7.8pt} 
\begin{tabular}{l|c|cccccccc}
\hline \toprule
Dataset          & Pub.    & \multicolumn{4}{c|}{CIFAR-10-LT}               & \multicolumn{4}{c}{CIFAR-100-LT} \\ \hline
Backbone Net     &\multicolumn{1}{c|}{-}  & \multicolumn{8}{c}{ResNet-32}                                 \\ \hline
Imbalance factor &\multicolumn{1}{c|}{-}  & 200  & 100  & 50   & \multicolumn{1}{c|}{10}   & 200    & 100    & 50    & 10  \\ \hline
Cross Entropy    &\multicolumn{1}{c|}{-}  & 65.6 & 70.3 & 74.8 & \multicolumn{1}{c|}{86.3} & 34.8   & 38.2   & 43.8   & 55.7  \\
BBN \cite{paper50}             & CVPR 2020    &\multicolumn{1}{c}{-}      & 79.8 & 82.1 & \multicolumn{1}{c|}{88.3} &\multicolumn{1}{c}{-}        & 42.5   & 47.0   & 59.1  \\
UniMix \cite{paper41}          & NeurIPS 2021 & 78.5 & 82.8 & 84.3 & \multicolumn{1}{c|}{89.7} & 42.1   & 45.5   & 51.1   & 61.3  \\
MetaSAug \cite{paper18}          & CVPR 2021    & 76.8 & 80.5 & 84.0 & \multicolumn{1}{c|}{89.4} & 39.9   & 46.8   & 51.9   & 61.7  \\
MiSLAS \cite{paper49}          & CVPR 2021    &\multicolumn{1}{c}{-}      & 82.1 &85.7 & \multicolumn{1}{c|}{90.0} &\multicolumn{1}{c}{-}        & 47.0   & 52.3   & 63.2  \\ 

CDB-W-CE \cite{paper29}        & IJCV 2022    &\multicolumn{1}{c}{-}      & \multicolumn{1}{c}{-}  & \multicolumn{1}{c}{-}  &\multicolumn{1}{c|}{-}  &\multicolumn{1}{c}{-}        & 42.6   & \multicolumn{1}{c}{-}    & 58.7  \\ 

GCL \cite{paper17}               & CVPR 2022    & 79.0 & 82.7 & 85.5 & \multicolumn{1}{c|}{-}     &44.9   & 48.7   & 53.6   &\multicolumn{1}{c}{-}       \\ 
RIDE + CR  \cite{paper52}               & CVPR 2023    & \multicolumn{1}{c}{-}  & \multicolumn{1}{c}{-}  & \multicolumn{1}{c}{-} & \multicolumn{1}{c|}{-}     &\multicolumn{1}{c}{-}   & 50.7   & \textbf{54.3}   &61.4       \\ \hline

OFA \cite{paper5}             & ECCV 2020    & 75.5 & 82.0 & 84.4 & \multicolumn{1}{c|}{\underline{\textbf{91.2}}} & 41.4   & 48.5   & 52.1   & \underline{\textbf{65.3}}  \\
M2m \cite{paper14}           & CVPR 2020    &\multicolumn{1}{c}{-}      & 78.3 &\multicolumn{1}{c}{-}      & \multicolumn{1}{c|}{87.9} &\multicolumn{1}{c}{-}        & 42.9   &\multicolumn{1}{c}{-}        & 58.2  \\
RSG \cite{paper34}             & CVPR 2021    &\multicolumn{1}{c}{-}      & 79.6 & 82.8 &\multicolumn{1}{c|}{-}     &\multicolumn{1}{c}{-}        & 44.6   & 48.5   &\multicolumn{1}{c}{-}       \\
CMO \cite{paper26}             & CVPR 2022    &\multicolumn{1}{c}{-}      &\multicolumn{1}{c}{-}      &\multicolumn{1}{c}{-}      & \multicolumn{1}{c|}{-}     &\multicolumn{1}{c}{-}        & 50.0   & 53.0   & 60.2  \\ 
FDC \cite{paper55}  & TMM 2023    &\textbf{79.7}      &\textbf{83.4}     & \underline{\textbf{86.5}}      &90.6     &\textbf{45.8}       & \textbf{50.6}   & 54.1   & 61.3  \\ \hline
FUR-Decoupled         &\multicolumn{1}{c|}{-}         &79.6 &\textbf{83.4} & 86.1 & \multicolumn{1}{c|}{90.7} & \textbf{45.8}   & \textbf{50.7}   & \textbf{53.9}   &61.4  \\
FUR       &\multicolumn{1}{c|}{-}         & \underline{\textbf{79.8}} & \underline{\textbf{83.7}} & \textbf{86.2} & \multicolumn{1}{c|}{\textbf{90.9}} & \underline{\textbf{46.2}}   & \underline{\textbf{50.9}}   & \underline{\textbf{54.1}}   & \textbf{61.8}  \\   \bottomrule \hline
\end{tabular}
\end{table*}

\begin{table}[h]
\caption{Details of the experimental setup. The $100+50+50$ in Epoch indicates the first phase, the second phase, and the third phase are trained for $100$, $50$, and $50$ epochs, respectively.}
\label{table2}
\vskip 0.05in
\centering  
\renewcommand\arraystretch{1.4}
\setlength{\tabcolsep}{0.2pt} 
\begin{tabular}{ccccc}
\hline \toprule
\multicolumn{2}{c|}{Dataset}                                           & \multicolumn{1}{c|}{CIFAR-10/100-LT} & \multicolumn{1}{c|}{ImageNet-LT} & iNaturalist 2018 \\ \hline
\multicolumn{2}{c|}{Backbone}                                          & \multicolumn{1}{c|}{ResNet-32}       & \multicolumn{1}{c|}{ResNeXt-50}  & ResNet-50        \\ \hline
\multicolumn{2}{c|}{Epoch}                                             & \multicolumn{1}{c|}{100+50+50}       & \multicolumn{1}{c|}{100+50+50}   & 100+50+50        \\ \hline
\multicolumn{5}{c}{Optimizer: SGD}                                                                                                                                  \\ \hline
\multicolumn{2}{c|}{Mm}                                                & \multicolumn{3}{c}{0.9}                                                                    \\ \hline
\multicolumn{1}{c|}{\multirow{3}{*}{LR}} & \multicolumn{1}{c|}{Phase1} & \multicolumn{1}{c|}{0.05}            & \multicolumn{1}{c|}{0.1}         & 0.1              \\ \cmidrule(lr){2-5} 
\multicolumn{1}{c|}{}                    & \multicolumn{1}{c|}{Phase2} & \multicolumn{1}{c|}{0.001}           & \multicolumn{1}{c|}{0.001}       & 0.001            \\ \cmidrule(lr){2-5} 
\multicolumn{1}{c|}{}                    & \multicolumn{1}{c|}{Phase3} & \multicolumn{1}{c|}{0.001}           & \multicolumn{1}{c|}{0.001}       & 0.001            \\ \hline
\multicolumn{2}{c|}{LR decay}                                          & \multicolumn{1}{c|}{Cosine}          & \multicolumn{1}{c|}{Linear}      & Linear           \\ \hline
\multicolumn{2}{c|}{Batch size}                                        & \multicolumn{1}{c|}{128}             & \multicolumn{1}{c|}{256}         & 512              \\ \hline
\multicolumn{2}{c|}{Warm-up}                                           & \multicolumn{1}{c|}{\Checkmark}                & \multicolumn{1}{c|}{\XSolidBrush}            &                 \XSolidBrush  \\  \bottomrule  \hline
\end{tabular}
\end{table}

\begin{table*}[t]
\caption{Top-1 accuracy (\%) of ResNext-50 \cite{paper20} on ImageNet-LT and Top-1 accuracy (\%) of ResNet-50 \cite{paper21} on iNaturalist 2018 for classification. The best and the second-best results are shown in \underline{\textbf{underline bold}} and \textbf{bold}, respectively. FUR-Decoupled indicates a FUR with decoupled training, and FUR Default indicates a FUR with three-stage training scheme.}
\label{table3}
\centering  
\renewcommand\arraystretch{1}
\setlength{\tabcolsep}{5.7pt} 
\begin{tabular}{l|c|cccc|cccc}
\hline \toprule
\multirow{3}{*}{Methods} & \multirow{3}{*}{Pub.} & \multicolumn{4}{c|}{ImageNet-LT} & \multicolumn{4}{c}{iNaturalist 2018} \\  \cmidrule(lr){3-10}
                         &                       & \multicolumn{4}{c|}{ResNext-50}  & \multicolumn{4}{c}{ResNet-50}    \\ \cmidrule(lr){3-10}
                         &                       & Head  & Middle  & Tail & Overall & Head  & Middle  & Tail & Overall \\ \hline
BBN \cite{paper50}                   & CVPR 2020              & 43.3  & 45.9    & \underline{\textbf{43.7}} & 44.7    & 49.4  & 70.8    & 65.3 & 66.3    \\
DisAlign \cite{paper45}               & CVPR 2021                  & 59.9  & 49.9    & 31.8 & 52.9    & 68.0  & 71.3    & 69.4 & 70.2    \\
UniMix \cite{paper41}                  & NeurIPS 2021               &\multicolumn{1}{c}{-}       &\multicolumn{1}{c}{-}         &\multicolumn{1}{c}{-}      & 48.4    &\multicolumn{1}{c}{-}       &\multicolumn{1}{c}{-}         &\multicolumn{1}{c}{-}      & 69.2    \\
MetaSAug \cite{paper18}                & CVPR 2021                  &\multicolumn{1}{c}{-}       &\multicolumn{1}{c}{-}         &\multicolumn{1}{c}{-}      & 47.3    &\multicolumn{1}{c}{-}       &\multicolumn{1}{c}{-}         &\multicolumn{1}{c}{-}      & 68.7    \\
MiSLAS \cite{paper49}                 & CVPR 2021                  & \textbf{65.3}  & \textbf{50.6}    & 33.0 & 53.4    &\textbf{73.2}  & \textbf{72.4}    & 70.4 & 71.6    \\

CDB-W-CE \cite{paper29}                 & IJCV 2022                  & \multicolumn{1}{c}{-}  &\multicolumn{1}{c}{-}     & \multicolumn{1}{c}{-}  & 38.5    & \multicolumn{1}{c}{-}   & \multicolumn{1}{c}{-}    & \multicolumn{1}{c}{-}  & \multicolumn{1}{c}{-}   \\ 
GCL \cite{paper17}                   & CVPR 2022               &\multicolumn{1}{c}{-}       &\multicolumn{1}{c}{-}         &\multicolumn{1}{c}{-}      & \textbf{54.9}    &\multicolumn{1}{c}{-}       &\multicolumn{1}{c}{-}         &\multicolumn{1}{c}{-}      & \textbf{72.0}    \\ 
DSB-LADE \cite{paper53}                   & ICLR 2023               &62.6      &50.4         &33.6      &53.2    &72.3       &70.7         &65.8      &70.5 \\
LDAM + CR \cite{paper52}                   & CVPR 2023               &60.8      &50.3         &33.6      &52.4    &69.3       &66.7         &61.9      &65.7   \\ \hline

OFA \cite{paper5}                   & ECCV 2020          & 47.3  & 31.6    & 14.7 & 35.2    &\multicolumn{1}{c}{-}       &\multicolumn{1}{c}{-}         &\multicolumn{1}{c}{-}      & 65.9    \\
RSG \cite{paper34}                    & CVPR 2021             & 63.2  & 48.2    & 32.3 & 51.8    &\multicolumn{1}{c}{-}       &\multicolumn{1}{c}{-}         &\multicolumn{1}{c}{-}      & 70.2    \\
GistNet \cite{paper20}              & ICCV 2021              & 52.8  & 39.8    & 21.7 & 42.2    &\multicolumn{1}{c}{-}       &\multicolumn{1}{c}{-}         &\multicolumn{1}{c}{-}      & 70.8    \\
BS + CMO \cite{paper26}                & CVPR 2022                & 62.0  & 49.1    & 36.7 & 52.3    & 68.8  & 70.0    & \textbf{72.3} & 70.9    \\ \hline

FUR-Decoupled                &\multicolumn{1}{c|}{-}  &\textbf{65.1}  &\textbf{51.6}    & \textbf{38.3} &\textbf{55.2}    & \textbf{73.4}  & \textbf{72.5}    & \underline{\textbf{73.7}} &\textbf{72.4}  \\

FUR              &\multicolumn{1}{c|}{-}  & \underline{\textbf{65.4}}  & \underline{\textbf{52.2}}    & 37.8 & \underline{\textbf{55.5}}    & \underline{\textbf{73.6}}  & \underline{\textbf{72.9}}    &\textbf{73.1} & \underline{\textbf{72.6}}    \\  \bottomrule \hline
\end{tabular}
\end{table*}

\subsection{Implementation Details}\label{sec5.2}

Following the accepted settings \cite{paper6,paper45,paper49}, the batch sizes on ImageNet-LT and iNaturalist 2018 were taken to be $256$ and $512$, respectively. For a fair comparison, we are consistent with OFA \cite{paper5} and take $N_A$ to be $3$, so $N_T$ is $32$ and $64$ on ImageNet-LT and iNaturalist 2018, respectively. More details of the experimental setup are listed in Table \ref{table2}. The experimental setup on OIA-ODIR will be presented separately in Section \ref{sec5.7}. We trained models on CIFAR-10-LT, CIFAR-100-LT and OIA-ODIR using a single NVIDIA 2080Ti GPU and on ImageNet-LT and iNaturalist 2018 using $4$ NVIDIA 2080Ti GPUs.

\subsection{Comparative Methods}\label{sec5.3}

We train the proposed Feature Uncertainty Representation (FUR) employing decoupled training and three-stage training schemes, respectively. The FUR is then compared with classical and state-of-the-art long-tailed knowledge transfer methods, non-transfer data augmentation methods, and other state-of-the-art long-tailed recognition methods. The specific methods are classified as follows.

\begin{itemize}[]
\item \textbf{Classical and latest long-tailed knowledge transfer methods}, include OFA \cite{paper5}, M2m \cite{paper14}, RSG \cite{paper34}, GistNet \cite{paper20}, CMO \cite{paper26} and FDC \cite{paper55}.
\item \textbf{Other state-of-the-art methods}. They include the two-stage MiSLAS \cite{paper49}, DisAlign \cite{paper45}, BBN \cite{paper50} with two branches, and the non-transfer augmentation methods UniMix \cite{paper41}, MetaSAug \cite{paper18}, CDB \cite{paper29}, GCL \cite{paper17} DSB \cite{paper53} and CR \cite{paper52}.
\end{itemize}

\subsection{Results on CIFAR-10-LT and CIFAR-100-LT}\label{sec5.4}

The results on CIFAR-10-LT and CIFAR-100-LT are summarized in Table \ref{table1}, where our proposed method achieves optimal performance on six long-tailed CIFAR datasets and second-best results on the remaining two datasets. Our proposed FUR outperforms GCL by $\textbf{1\%}$ and $\textbf{2.2\%}$ on CIFAR-10-LT and CIFAR-100-LT with IF = $100$, respectively. On the CIFAR-100-LT with IF = $10$, FUR-Decoupled outperforms the combined CMO by $\textbf{1.6\%}$. FUR with a three stage training scheme outperforms FUR-Decoupled on all datasets, which we will analyze in detail in the next part of the experiment.

Compared to CMO, which randomly pastes the image foreground of the tail class onto the background of the head class image, our proposed FUR relies on the observed prior knowledge to recover the underlying distribution of the tail class. GCL constructs the same ``feature cloud" for each feature of the tail class to adjust the model logit, without taking into account the differences in domain characteristics between classes. As a result, FUR outperforms similar methods on multiple datasets.

\subsection{Results on ImageNet-LT and iNaturalist 2018}\label{sec5.5}

We report in Table \ref{table3} not only the overall performance of FUR and FUR-Decoupled on ImageNet-LT and iNaturalist 2018 but also additionally add the performance on three subsets of these two datasets, Head (more than $100$ images), Middle ($20\sim100$ images), and Tail (less than $20$ images). Compared to other methods, FUR shows the state-of-the-art overall performance on both ImageNet-LT and iNaturalist 2018.

We argue that although the bias of the classifier is mitigated after decoupled training, it is ignored whether the feature sub-network can adapt to the new decision boundaries, which leads to a trade-off in the performance of the head classes. Therefore we add a third stage to fine-tune the feature extractor to adapt it to the latest decision boundaries. FUR-Decoupled outperforms the transfer-based CMO by $\textbf{3.4\%}$ and $\textbf{4.8\%}$, respectively, on the Head subset of the two large-scale long-tailed datasets, benefiting from the fact that FUR relies on prior knowledge rather than randomly recovering the tail class distribution. Although there is a slight degradation in tail class performance after the third stage, the overall performance of the model and the performance on the head subset are better than the decoupled trained model. Thus both the feature sub-network and the classifier need to be fine-tuned to rebalance the preferences of the model. In addition, the extraordinary performance of FUR-Decoupled on tail classes suggests that our method can recover the underlying distribution of tail classes more efficiently.

\subsection{Results on OIA-ODIR}\label{sec5.7}

\begin{figure}[t]
	\centering
	\begin{minipage}{0.493\linewidth}
		\centering
		\includegraphics[width=0.99\linewidth]{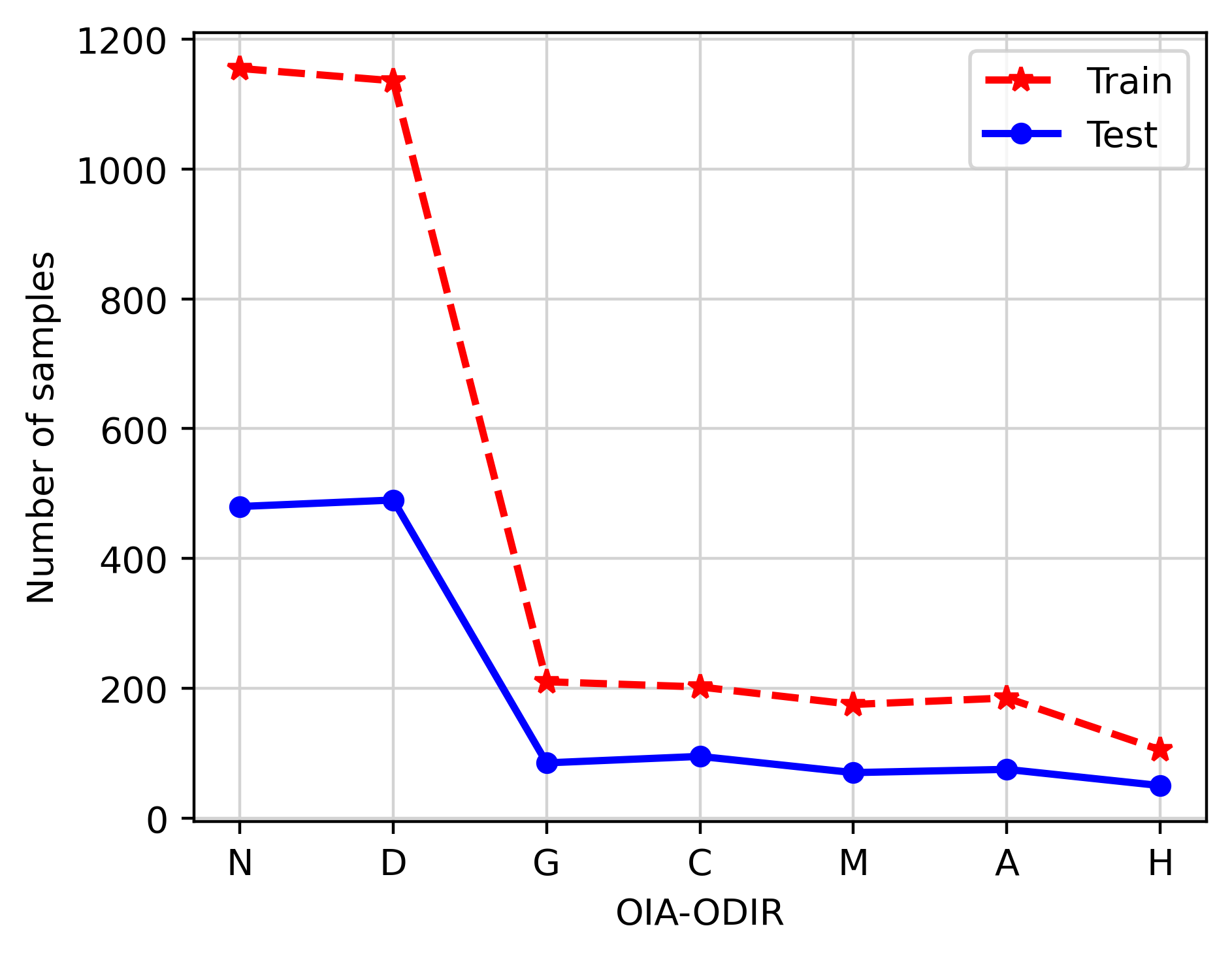}
	\end{minipage}
	\begin{minipage}{0.493\linewidth}
		\centering
		\includegraphics[width=0.94\linewidth]{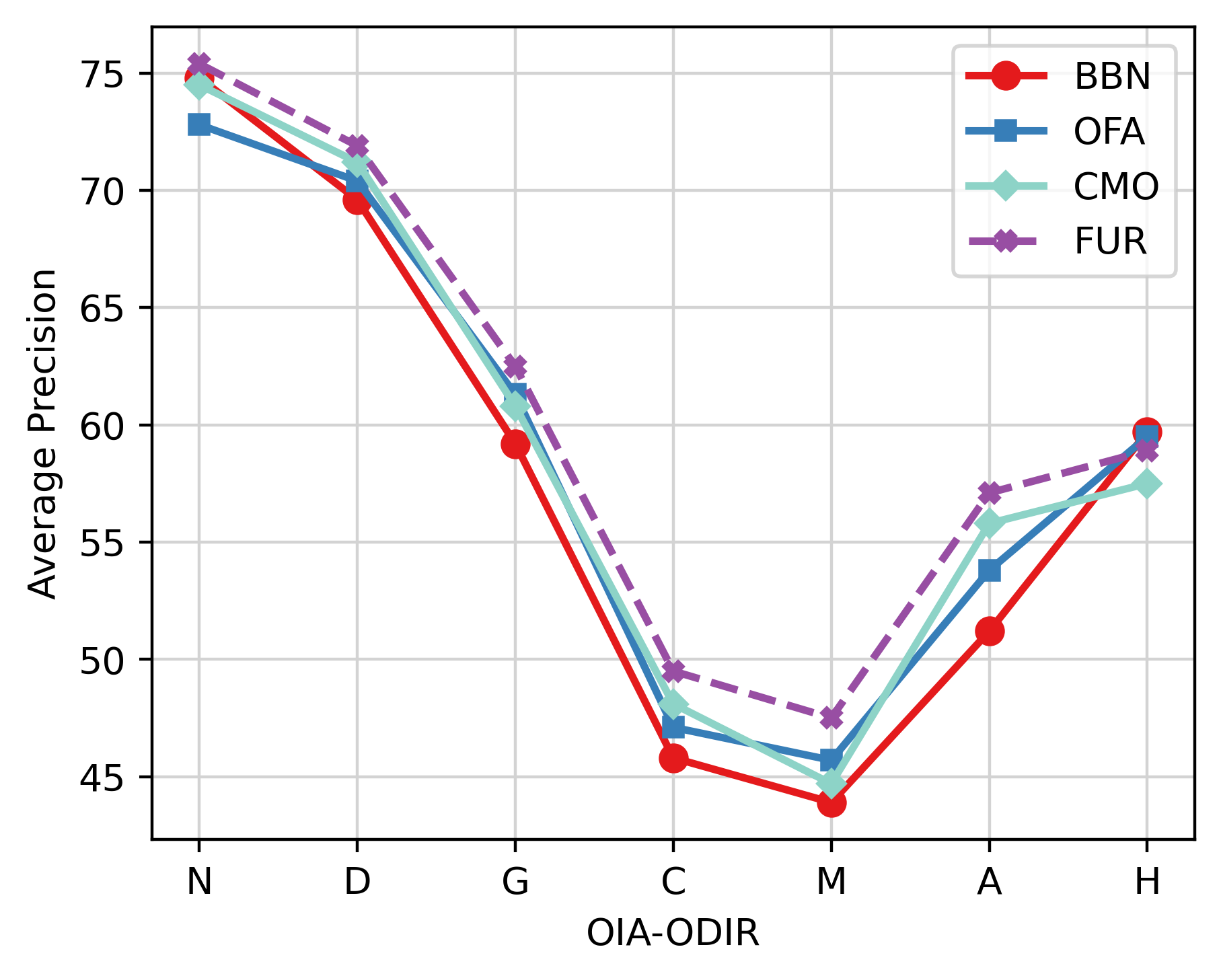}
	\end{minipage}
\caption{The left figure shows the number of training and testing samples for each class in OIA-ODIR dataset. The right figure presents a comparison of various methods.}
\label{figa1}
\end{figure}

The OIA-ODIR dataset contains a total of $10,000$ fundus images in $8$ classes. The eight classes are Normal (N), hypertensive retinopathy (D), glaucoma (G), cataract (C), agerelated macular degeneration (A), hypertension complication (H), pathologic myopia (M), other disease/abnormality (O). Considering that O usually appears together with other diseases, to reduce ambiguity, we adopt the data splitting scheme of \cite{paper53}, using only the data of the first $7$ classes, and the number of training samples and test samples for each class is shown in  Figure \ref{figa1}.

We used ResNet-50 as the backbone network. An adam optimizer with a learning rate of $0.1$ (linear decay), a momentum of $0.9$, and a weight decay factor of $0.005$ was adopted to train all networks. In keeping with \cite{paper53}, average precision (AP) was used as the performance metric of the model.
We implemented BBN, OFA, and CMO on the OIA-ODIR dataset and compared them with our method. The experimental results (Figure \ref{figa1}) indicate that FUR outperforms the other three methods.

\subsection{Visualization Analysis}\label{sec5.6}

\begin{figure}[h]
\vskip -0.1in
\centering
\includegraphics[width=3in]{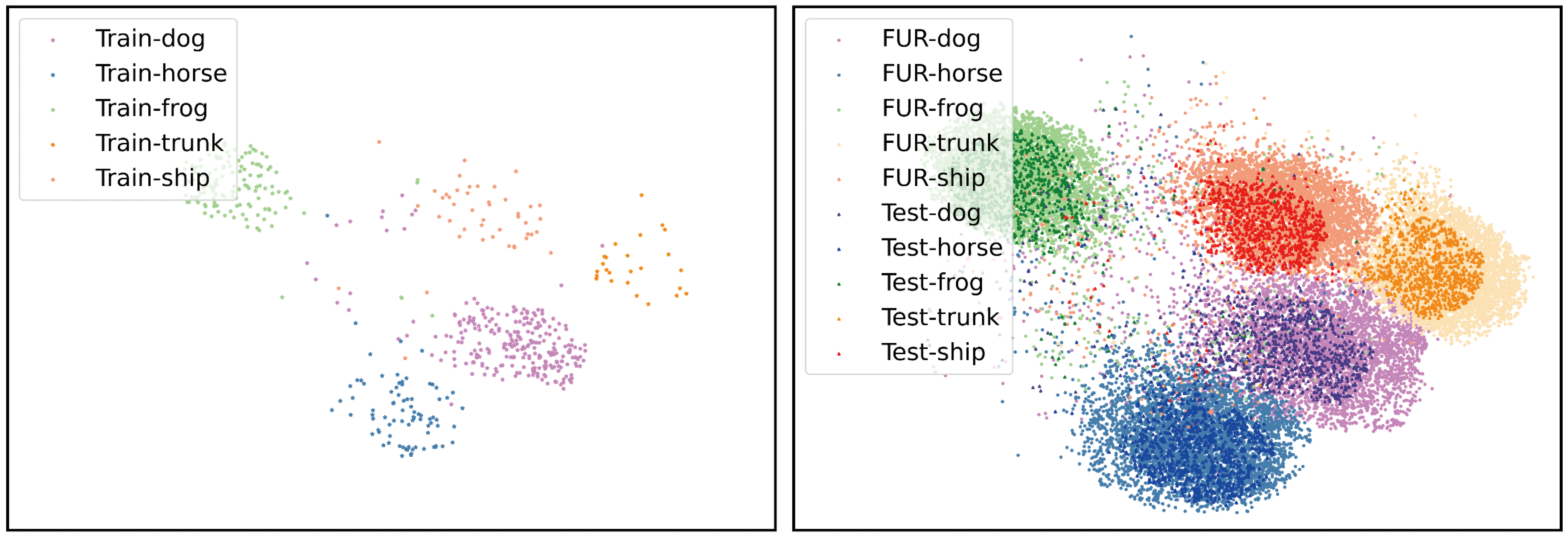}
\vskip 0.05in
\caption{Visualization of tail class feature embedding from CIFAR-10-LT with an imbalance factor of $200$.}
\label{fig5}
\vskip -0.4in
\end{figure}

To clearly demonstrate that FUR can excel in the recovery of the underlying distribution of tail classes, we visualized the tail features of CIFAR-10-LT via t SNE. As shown in Figure \ref{fig5}, the training distribution after augmentation with FUR can cover the test distribution well. The above results further show that our proposed method efficiently recovers the distribution of tail classes. This result further indicates that our proposed method accurately recovers the underlying distribution of the tail classes, allowing the model to perform better on the test set outside the training domain.

\subsection{More Analysis and Discussion}\label{sec5.8}

After the training in the third stage, the shape of the distribution of the head class may change. In this case, if we match the most similar head class for the tail class, it may cause the most similar head class to change. Then is there a need for further feature augmentation for the tail class? We do not think it is needed for the following reasons.

In the second stage, we enhanced the features of the tail category to make the model learn as much information as possible about the true distribution of the tail category. In the third stage, we used a smaller learning rate and trained with long-tailed data, which made the model more focused on learning about the head category. Therefore, after fine-tuning in the third stage, even if there is a slight change in the representation of the head category, the representation of the tail category by the model is almost unchanged, and knowledge enhancement for the tail category has been completed in the second stage. That is to say, even after training in the third stage, the shift of the tail class distribution is still small enough that features extracted by a model trained in three stages can still be well covered by augmented distributions generated by FUR in stage two. Figure \ref{fig5} demonstrates that the data distribution generated by FUR in the second stage can effectively cover the testing distribution. Therefore, we believe that the model has already achieved the goal of learning more about the underlying distribution of the tail classes, and the superior performance on the tail subsets of ImageNet-LT and iNaturalist 2018 further confirms the superiority of our method.
Additionally, adding extra training to continue adjusting the feature augmentation for the tail classes would increase the time cost of the training.

\begin{figure}[t]
	\centering
	\begin{minipage}{0.493\linewidth}
		\centering
		\includegraphics[width=0.99\linewidth]{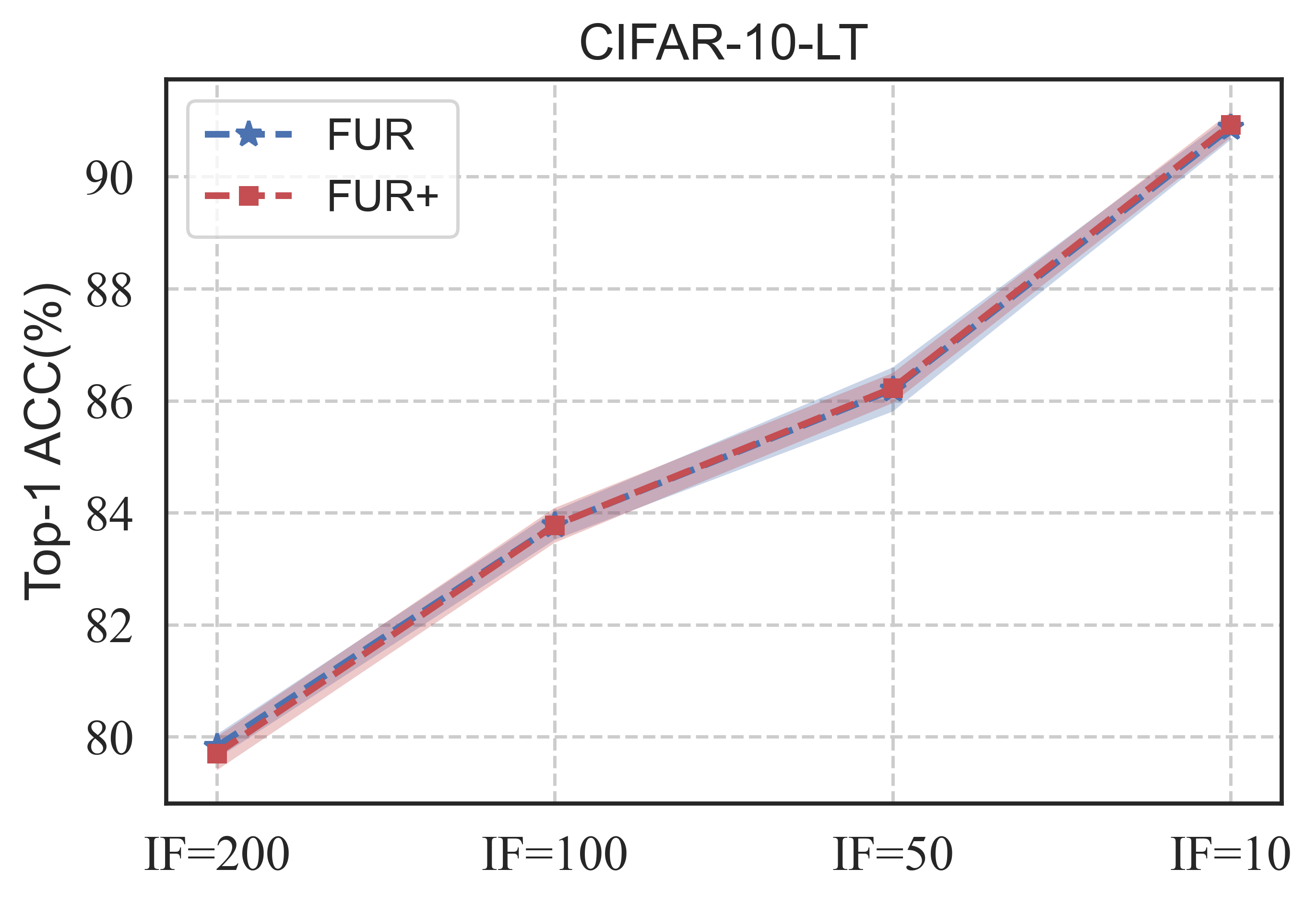}
	\end{minipage}
	\begin{minipage}{0.493\linewidth}
		\centering
		\includegraphics[width=0.96\linewidth]{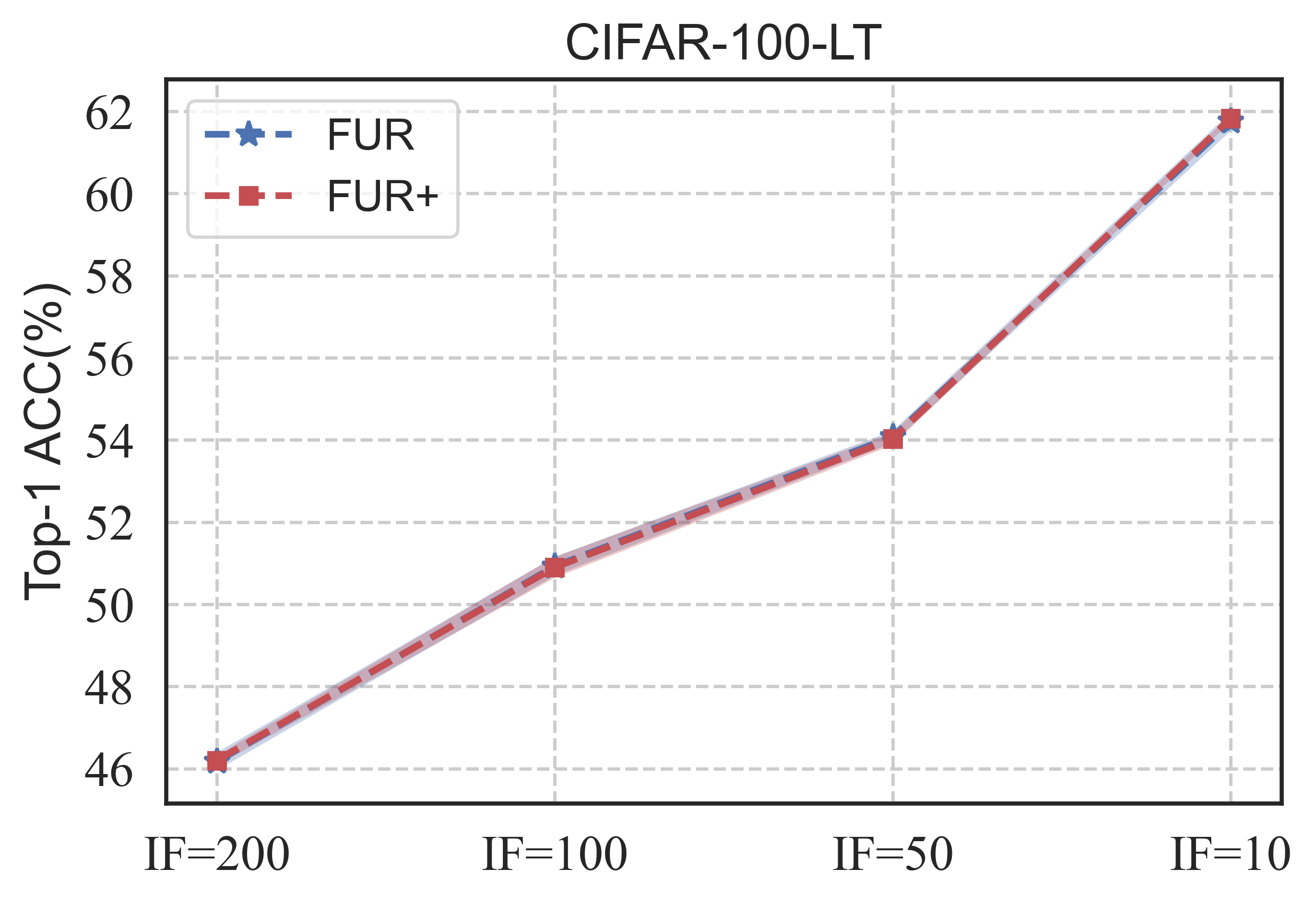}
	\end{minipage}
\caption{FUR+ denotes the addition of a fourth training phase to retune the feature augmentation of the tail class, and FUR denotes the method proposed in this paper.}
\label{figa2}
\end{figure}

We added training stage to adjust the feature enhancement of the tail class on CIFAR-10-LT and CIFAR-100-LT and compared it with the method proposed in this paper. The parameters of the fourth stage are consistent with those of the third stage. As shown in the Figure \ref{figa2}, the performance of the two methods is almost the same, and the confidence intervals of the five groups of results completely overlap. For the above reasons, we finally proposed a three-stage training strategy.

\section{Conclusion}\label{sec6}
In this work, We discovered four fundamental phenomena regarding the relationship between the geometry of feature distributions, which provide the theoretical and experimental basis for subsequent studies of class imbalance. Inspired by the four phenomena, we propose feature uncertainty representation (FUR) with geometric information for recovering the true distribution of tail classes. After three stages of training, the experimental results show that our proposed method greatly improves the performance of the tail class compared to other methods and ensures the superior performance of the head class at the same time.

\section{Data Availability Statements}\label{sec7}
All datasets used in this study are open-access and have been cited in the paper.

\begin{appendices}

\section*{Appendix A}\label{secA}

\begin{figure}[h]
\vskip -0.1in
\centering
\includegraphics[width=3in]{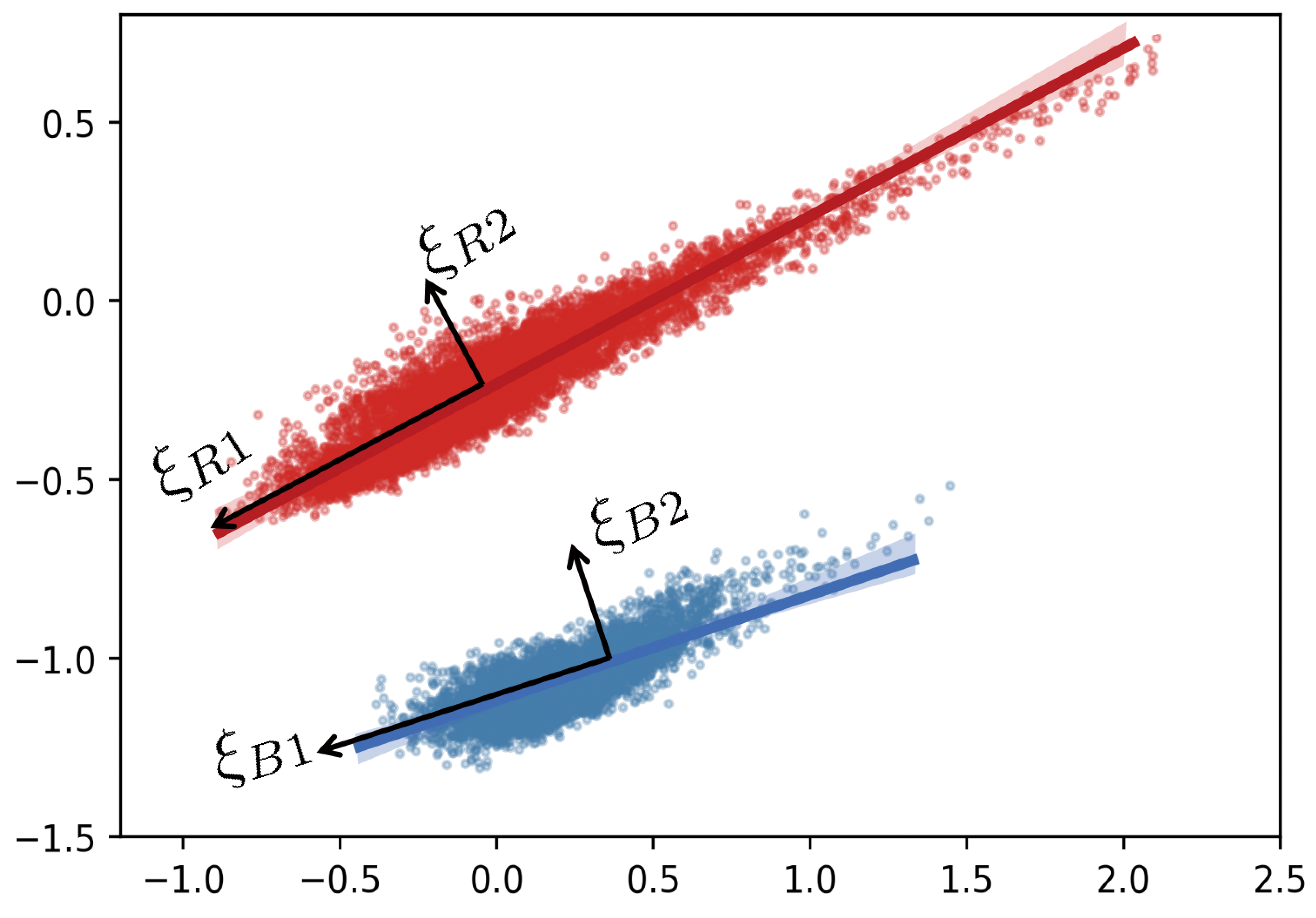}
\vskip 0.05in
\caption{Two distributions with similar geometry.}
\label{fig6}
\vskip -0.1in
\end{figure}

\begin{figure}[h]
\vskip -0.2in
\centering
\includegraphics[width=3in]{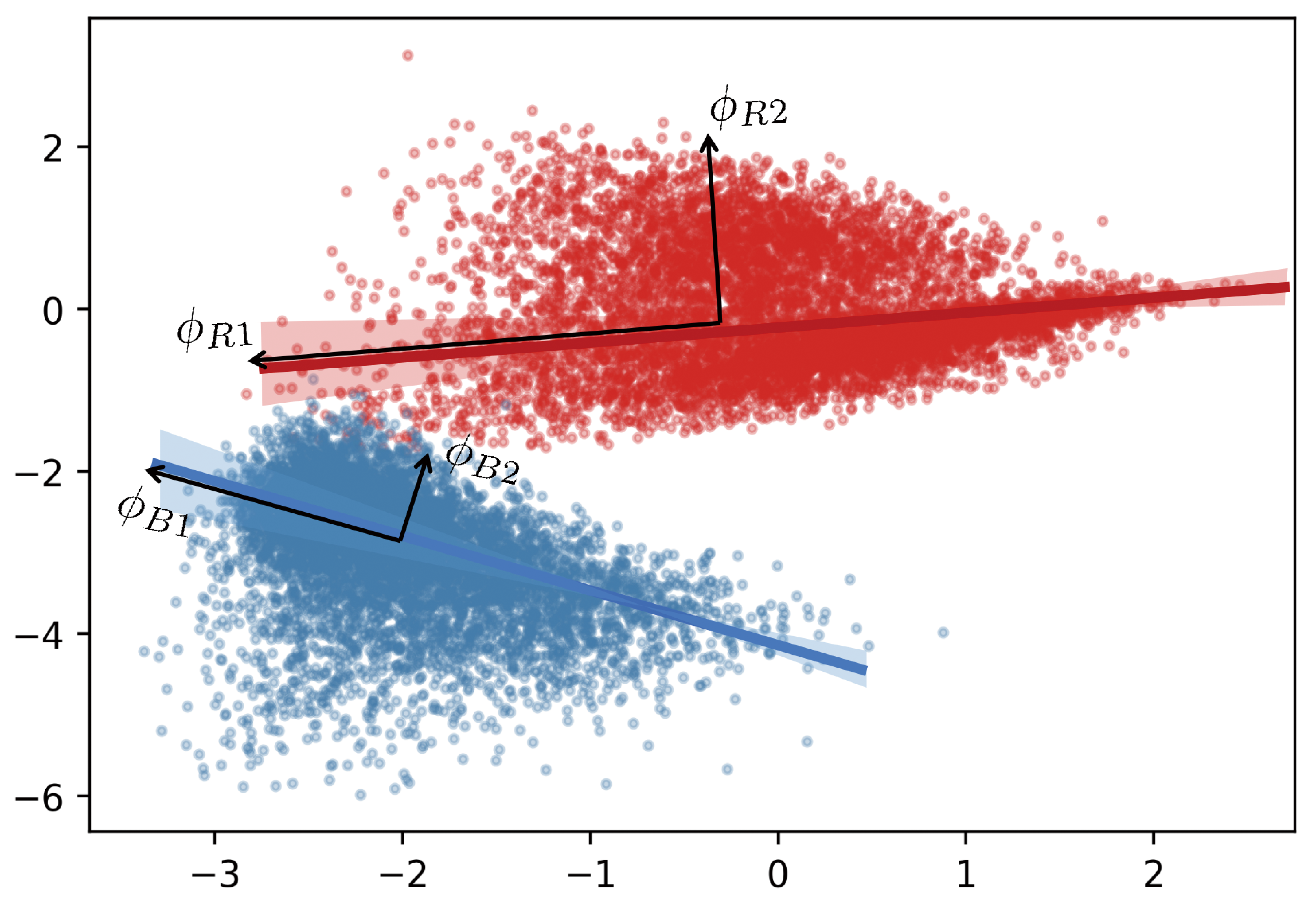}
\vskip 0.05in
\caption{Two distributions with low geometry similarity.}
\label{fig7}
\vskip -0.1in
\end{figure}

To facilitate the analysis and understanding of the geometry of the feature distributions and the similarity between the geometry, four two-dimensional distributions were generated and plotted in Figure \ref{fig6} and Figure \ref{fig7}. The geometry of the feature distribution is first introduced. As shown in Figure \ref{fig6}, the direction $\xi_{R1}$ with the largest variance and the direction $\xi_{R2}$ with the largest variance in the direction orthogonal to $\xi_{R1}$ are selected. It can be seen that the geometry and location of the distribution can be anchored by $\xi_{R1}$, $\xi_{R2}$ and the center of the distribution. It is important to note that in this work, we only focus on the shape of the distribution and ignore the location of the distribution. Moreover, if the projection is done along these two directions, the information of the distribution is preserved to the maximum extent.

Observing Figure \ref{fig6}, we can notice that the geometry of the red and blue distributions are more similar because their covariances are similar, i.e., the pattern of variation of the vertical axis with the horizontal axis is similar. The direction of the maximum variance mainly determines the shape of the distribution, and the direction of the second largest variance also plays a role in the shape of the distribution. Obviously, the geometry of the two distributions in Figure \ref{fig6} is more similar than in Figure \ref{fig7}.

\section*{Appendix B}\label{secB}

The CIFAR-10 dataset consists of $60,000$ images with size $32\times 32$ from $10$ classes, each class contains $6,000$ images, of which $5,000$ images are used for training and $1,000$ images are used for testing. Fashion MNIST has ten classes, each containing $6000$ training images and $1000$ test images, each with a size of $28\times 28$.

\begin{figure}[h]
\vskip -0.2in
\centering
\includegraphics[width=3in]{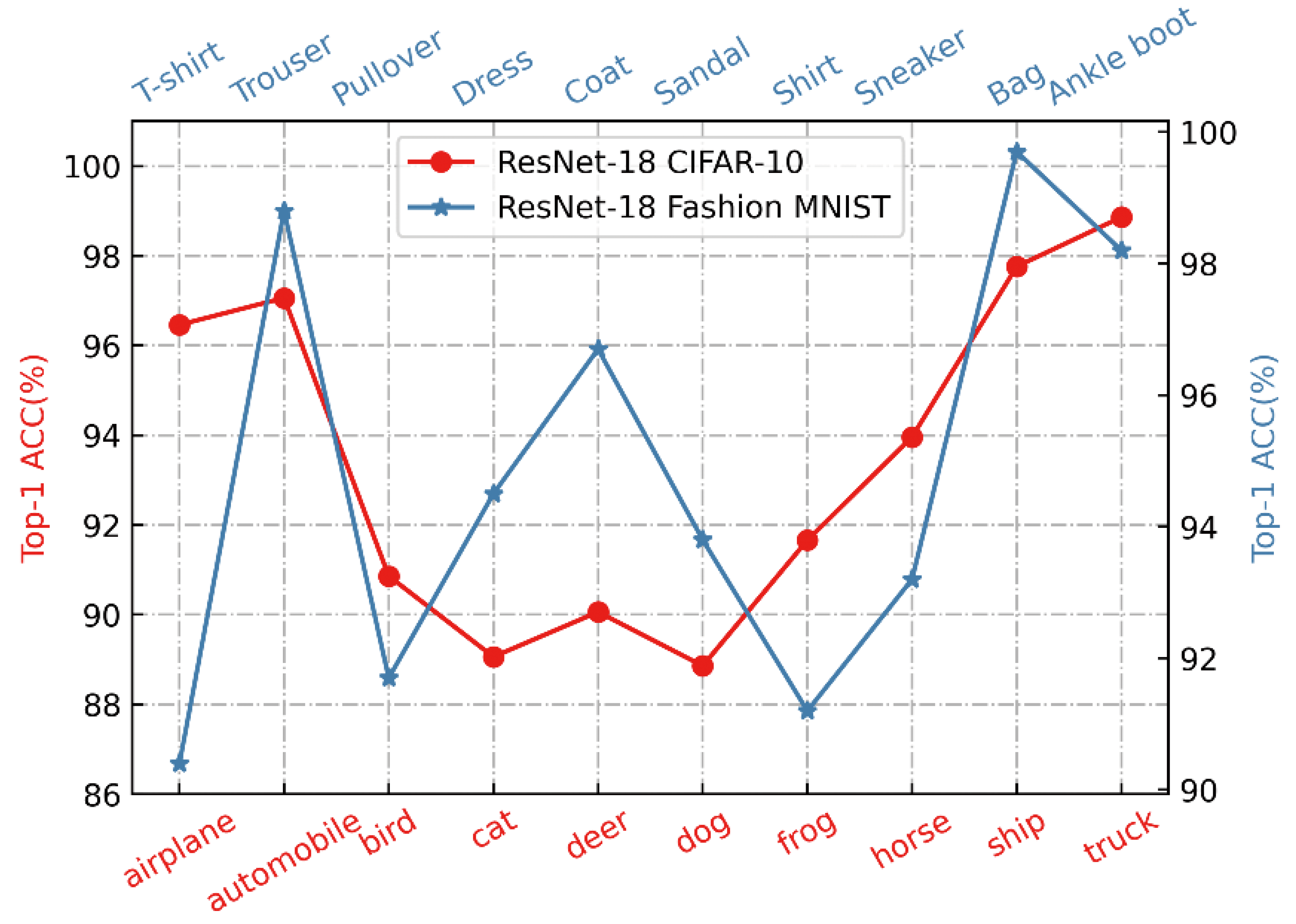}
\vskip 0.05in
\caption{Class accuracy of ResNet-18 on Fashion MNIST and CIFAR-10. The class indexes in Figure \ref{fig2} correspond to the ten numbers of MNIST. For Fashion MNIST and CIFAR-10, the class indexes (i.e., $1$ to $10$) correspond to the classes from left to right in the above figure, respectively.}
\label{fig8}
\vskip -0.1in
\end{figure}

To improve the generalization ability of the model and prevent the model from overfitting on the training set, we perform three data augmentation operations on the training set: random flip, random crop, and Cutout. Cutout keeps the model from being overly dependent on certain areas of the image by randomly masking out parts of the image. Considering the size of the image is small, the size $7\times 7$ convolutional kernel of ResNet-18 and the pooling operation tend to lose spatial information, so we remove the maximum pooling layer and adopt the size $3\times 3$ convolutional kernel instead of the size $7\times 7$ convolutional kernel.

We adopt SGD to optimize the model, set the batch size to $128$, and the initial learning rate to $0.1$. If the loss does not decrease after $10$ consecutive epochs, the learning rate becomes $0.5$ times of the original, and we train a total of $250$ epochs. ResNet-18 achieved an accuracy of $93.46\%$ on CIFAR-10 and $94.82\%$ on Fashion MNIST. The accuracy rates for each class are plotted in Figure \ref{fig8}.

\section*{Appendix C}\label{secC}

In the following, we derive the probability density function of the inner product of two random vectors. Without loss of generality, we set $x$ to be a $P$-dimensional random unit vector and fix $y$ to be a unit vector, i.e.
\begin{equation}
\begin{split}
x=(x_1,x_2,\dots,x_P), y=(1,0,\dots,0).
\nonumber
\end{split}
\end{equation}
The above equation satisfies $x_1^2+x_2^2+\cdots+x_P^2=1$. Using spherical transformations
\begin{equation}
\begin{split}
\left\{
\begin{array}{ll}
x_1=r\cos\varphi_1,  \\
x_2=r\sin\varphi_1\cos\varphi_2, \\
x_3=r\sin\varphi_1 \sin \varphi_2 \cos\varphi_3,   \\
\cdots \\
x_{n-1}=r\sin\varphi_1\sin\varphi_2\cdots\sin\varphi_{n-2}\cos\varphi_{n-1}, \\
x_n=r\sin\varphi_1\sin\varphi_2\cdots\sin\varphi_{n-2}\sin\varphi_{n-1},
\end{array}
\right.
\nonumber
\end{split}
\end{equation}
where
\begin{equation}
\begin{split}
\left\{
\begin{array}{ll}
0 \le r \le +\infty , \\
0 \le \varphi_1 \le \pi, \\
\cdots \\
0 \le \varphi_{n-2} \le \pi, \\
0 \le \varphi_{n-1} \le 2 \pi.
\end{array}
\right.
\nonumber
\end{split}
\end{equation}
The Jacobi determinant of the above transformation is
\begin{equation}
\begin{split}
J&=\frac{\partial (x_1,x_2,\dots,x_n)}{\partial (r,\varphi_1,\dots,\varphi_{n-1})}  \\
&=r^{n-1}\sin^{n-2}\varphi_1 \sin^{n-3}\varphi_2 \cdots \sin\varphi_{n-2}.
\nonumber
\end{split}
\end{equation}

Since $x$ is a unit vector, $r=1$. Notice that $\langle x,y \rangle = x_1=\cos \varphi_1$, so $\cos \langle x,y \rangle = \cos \varphi_1$. According to the geometric probability,
\begin{equation}
\begin{split}
&P_n(\varphi_1 \le \theta)= \\
&\frac{\int_{0}^{\theta}\sin^{n-2}\varphi_1d\varphi_1 \cdots \int_{0}^{\pi}\sin\varphi_{n-2}d\varphi_{n-2}\int_{0}^{2\pi}d\varphi_{n-1}}{\int_{0}^{\pi}\sin^{n-2}\varphi_1d\varphi_1\cdots \int_{0}^{\pi}\sin^2\varphi_{n-2}d\varphi_{n-2}\int_{0}^{2\pi}d\varphi_{n-1}}  \\
&=\frac{S_{n-1}\int_{0}^{\theta}\sin^{n-2}\varphi_1d\varphi_1 }{S_n} 
\nonumber
\end{split}
\end{equation}

Where $S_n$ denotes the surface area of the $n$-dimensional unit sphere. When $k$ is a positive integer,
\begin{equation}
\begin{split}
\int_{0}^{\pi}\sin^{k-1}\varphi d \varphi=2\int_{0}^{\frac{\pi}{2}}\sin^{k-1}\varphi d\varphi,  
\nonumber
\end{split}
\end{equation}
and because
\begin{equation}
\begin{split}
\int_{0}^{\frac{\pi}{2}}\sin^n\varphi d \varphi= \left\{
\begin{array}{ll}
\frac{(2m-1)!!}{(2m)!!}\cdot \frac{\pi}{2},n=2m \\
\frac{(2m)!!}{(2m+1)!!},n=2m+1
\end{array}
\right.
,
\nonumber
\end{split}
\end{equation}
the expression of $S_n$ is obtained. For convenience, we can use the $\Gamma$ function to unify the two, then
\begin{equation}
\begin{split}
S_n=\frac{2\pi^{\frac{n}{2}}}{\Gamma(\frac{n}{2})}.
\nonumber
\end{split}
\end{equation}
We can obtain
\begin{equation}
\begin{split}
P_n(\varphi_1\le \theta)=\frac{\Gamma(\frac{n}{2})}{\Gamma(\frac{n-1}{2}\sqrt{\pi} )}\int_{0}^{\theta}\sin^{n-2}\varphi_1 d \varphi_1. 
\nonumber
\end{split}
\end{equation}
Further, the probability density function of $\theta$ is calculated as
\begin{equation}
\begin{split}
f_n(\theta)&=\frac{d}{d\theta}P_n(\varphi_1 \le \theta)  \\
&= \frac{\Gamma(\frac{n}{2})}{\Gamma(\frac{n-1}{2}\sqrt{\pi} )} \sin^{n-2} \theta.
\nonumber
\end{split}
\end{equation}

\begin{figure}[t]
\centering
\includegraphics[width=3in]{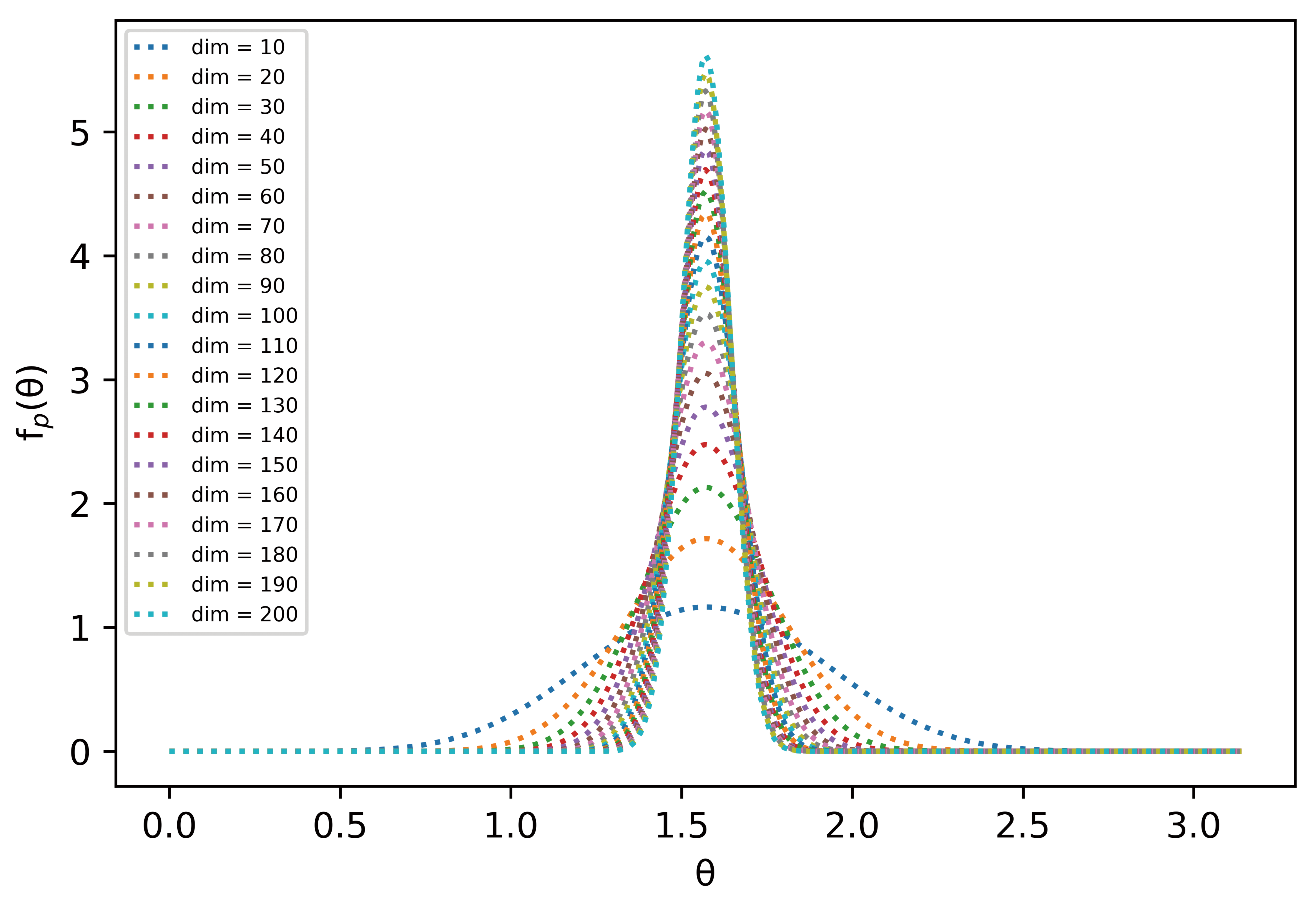}
\vskip 0.05in
\caption{The probability density function of the angle between two high-dimensional random vectors.}
\label{fig9}
\vskip -0.1in
\end{figure}

We plot the curve of the function $f_n(\theta)$ in Figure \ref{fig9}. It can be seen that the angle between the two random vectors tends to $\pi/2$ as the dimensionality increases.

\begin{figure}[h]
\vskip -0.1in
\centering
\includegraphics[width=3in]{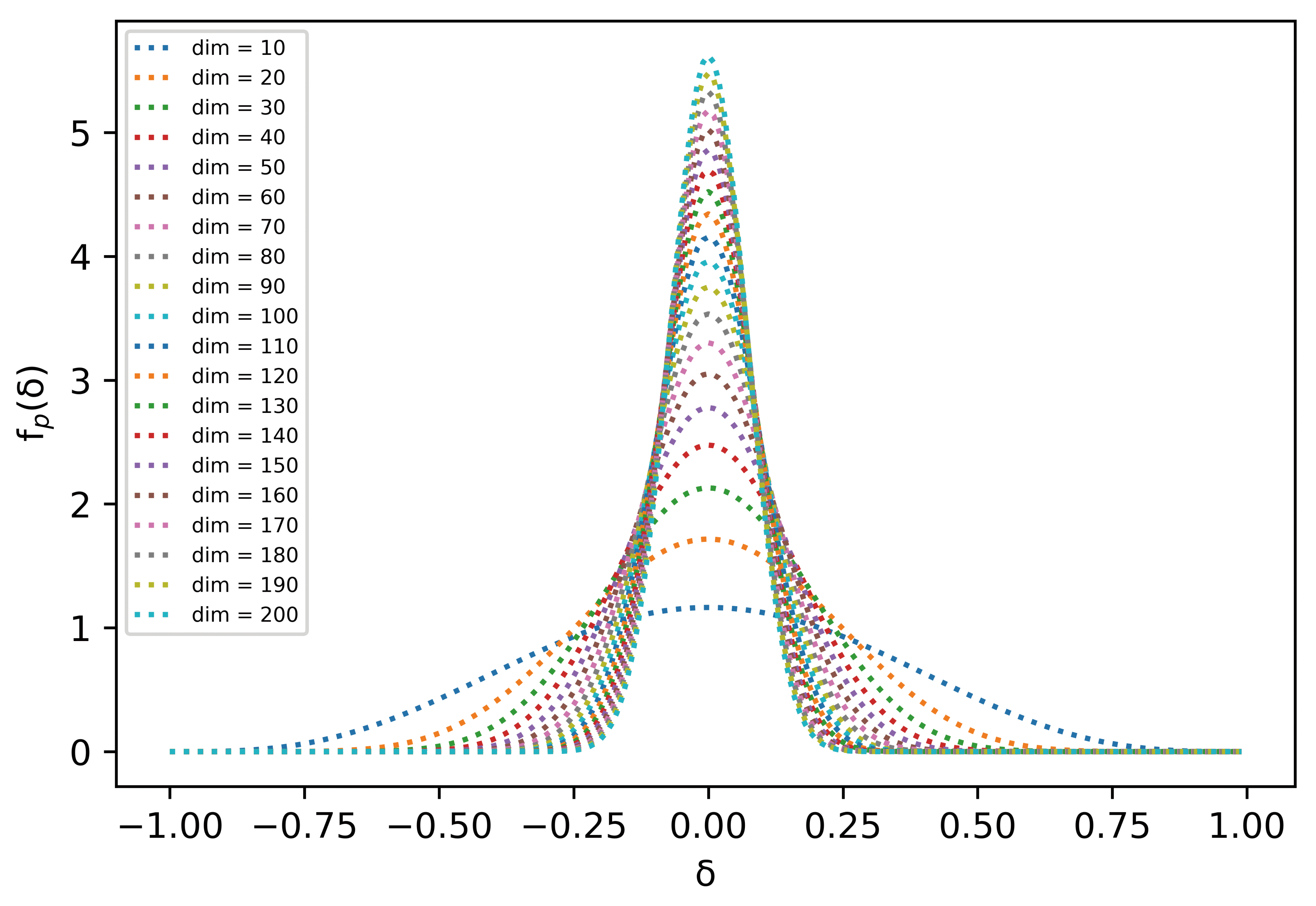}
\vskip 0.05in
\caption{The probability density function of the inner product between two high-dimensional random vectors.}
\label{fig10}
\vskip -0.3in
\end{figure}

Let $\delta=\cos \theta$. Then the probability density function of $\delta$ is
\begin{equation}
\begin{split}
f_n(\delta)=\frac{\Gamma(\frac{n}{2})}{\Gamma(\frac{n-1}{2}\sqrt{\pi} )}(1-\delta^2)^{\frac{n-3}{2}}.
\nonumber
\end{split}
\end{equation}
In the main text, the dimensionality of the feature vector is denoted by $P$. We plot the curve of the function $f_n(\delta)$ in Figure \ref{fig10}. It can be seen that the inner product between two high-dimensional random vectors tends to $0$ as the dimensionality increases, which means that the two random vectors tend to be orthogonal. The above results prove that our findings did not happen by chance and that the experimental phenomena we summarized are reliable.

\section*{Appendix D}\label{secD}

\begin{table*}[t]
\caption{Details of all classes in Figure \ref{fig3}a and Figure \ref{fig3}b.}
\label{table4}
\renewcommand\arraystretch{1}
\setlength{\tabcolsep}{5.1pt} 
\begin{tabular}{|l|l|l|l|l|l|l|l|l|l|}
\hline
         & Shirt    & Pullover & Dress   & Bag      & Trouser  & Sneaker & Ankle   & Sandal   & Coat     \\ \hline
T-shirt  & 3.63     & 2.56     & 1.27    & 0.69     & 1.15     & 0.93    & 1.28    & 1.16     & 1.02     \\ \hline
         & Dress    & Coat     & Bag     & Pullover & Shirt    & T-shirt & Ankle   & Sneaker  & Sandal   \\ \hline
Trouser  & 2.02     & 1.52     & 1.56    & 1.52     & 1.57     & 1.15    & 0.83    & 0.86     & 0.64     \\ \hline
         & Shirt    & Coat     & T-shirt & Dress    & Trouser  & Bag     & Sandal  & Ankle    & Sneaker  \\ \hline
Pullover & 3.42     & 2.62     & 2.56    & 1.20     & 1.52     & 1.52    & 1.48    & 0.71     & 0.36     \\ \hline
         & Coat     & Trouser  & Shirt   & T-shirt  & Pullover & Bag     & Ankle   & Sandal   & Sneaker  \\ \hline
Dress    & 2.58     & 2.02     & 1.97    & 1.27     & 1.20     & 1.32    & 1.24    & 0.58     & 0.48     \\ \hline
         & Pullover & Dress    & Shirt   & Bag      & Trouser  & Ankle   & Sneaker & T-shirt  & Sandal   \\ \hline
Coat     & 2.62     & 2.58     & 1.96    & 0.99     & 1.52     & 1.03    & 1.22    & 1.02     & 0.72     \\ \hline
         & Sneaker  & Ankle    & Bag     & Dress    & Shirt    & Trouser & T-shirt & Pullover & Coat     \\ \hline
Sandal   & 3.29     & 1.13     & 1.67    & 0.58     & 0.69     & 0.64    & 1.16    & 1.48     & 0.72     \\ \hline
         & T-shirt  & Pullover & Coat    & Dress    & Bag      & Trouser & Ankle   & Sandal   & Sneaker  \\ \hline
Shirt    & 3.63     & 3.42     & 1.96    & 1.97     & 1.36     & 1.57    & 1.41    & 0.69     & 0.41     \\ \hline
         & Sandal   & Ankle    & Coat    & T-shirt  & Trouser  & Bag     & Dress   & Shirt    & Pullover \\ \hline
Sneaker  & 3.29     & 2.65     & 1.22    & 1.07     & 0.86     & 1.24    & 0.48    & 0.41     & 0.36     \\ \hline
         & Sandal   & Dress    & Shirt   & Coat     & Pullover & T-shirt & Ankle   & Trouser  & Sneaker  \\ \hline
Bag      & 1.67     & 1.32     & 1.36    & 0.99     & 1.52     & 0.69    & 1.06    & 1.56     & 1.24     \\ \hline
         & Sneaker  & Sandal   & Bag     & Coat     & Dress    & Shirt   & T-shirt & Trouser  & Pullover \\ \hline
Ankle    & 2.65     & 1.13     & 1.06    & 1.03     & 1.24     & 1.41    & 1.28    & 0.83     & 0.71     \\ \hline
\end{tabular}

\vskip 0.2in
\renewcommand\arraystretch{1}
\setlength{\tabcolsep}{2.4pt} 
\begin{tabular}{|l|l|l|l|l|l|l|l|l|l|}
\hline
           & bird     & trunk      & ship     & cat   & horse    & deer     & automobile & frog       & dog        \\ \hline
airplane   & 3.43     & 3.05       & 1.70     & 1.20  & 0.81     & 1.94     & 0.81       & 1.64       & 1.43       \\ \hline
           & trunk    & ship       & airplane & forg  & cat      & horse    & bird       & dog        & deer       \\ \hline
automobile & 2.52     & 1.32       & 0.81     & 0.54  & 0.65     & 0.84     & 1.00       & 0.49       & 1.14       \\ \hline
           & airplane & deer       & cat      & dog   & frog     & horse    & ship       & automobile & trunk      \\ \hline
bird       & 3.43     & 2.83       & 2.28     & 1.47  & 1.39     & 1.45     & 1.18       & 1.00       & 0.56       \\ \hline
           & dog      & bird       & deer     & frog  & horse    & airplane & ship       & trunk      & automobile \\ \hline
cat        & 3.04     & 2.28       & 2.17     & 1.93  & 0.96     & 1.20     & 0.86       & 0.90       & 0.65       \\ \hline
           & bird     & cat        & horse    & dog   & frog     & airplane & ship       & automobile & truck      \\ \hline
deer       & 2.83     & 2.17       & 1.54     & 1.46  & 1.14     & 1.94     & 1.13       & 1.14       & 0.46       \\ \hline
           & cat      & horse      & bird     & deer  & frog     & airplane & truck      & ship       & automobile \\ \hline
dog        & 3.04     & 2.44       & 1.47     & 1.46  & 0.74     & 1.43     & 1.16       & 1.03       & 0.49       \\ \hline
           & cat      & bird       & deer     & dog   & airplane & ship     & horse      & automobile & truck      \\ \hline
frog       & 1.93     & 1.39       & 1.14     & 0.74  & 1.64     & 0.90     & 1.22       & 0.54       & 0.83       \\ \hline
           & dog      & deer       & cat      & bird  & airplane & truck    & frog       & ship       & automobile \\ \hline
horse      & 2.44     & 1.54       & 0.96     & 1.45  & 0.81     & 0.85     & 1.22       & 1.02       & 0.84       \\ \hline
           & airplane & automobile & truck    & cat   & bird     & frog     & deer       & horse      & dog        \\ \hline
ship       & 1.70     & 1.32       & 1.18     & 0.86  & 1.18     & 0.90     & 1.13       & 1.02       & 1.03       \\ \hline
           & airplane & automobile & ship     & horse & cat      & dog      & bird       & frog       & deer       \\ \hline
truck      & 3.05     & 2.52       & 1.18     & 0.85  & 0.90     & 1.16     & 0.56       & 0.83       & 0.46       \\ \hline
\end{tabular}
\end{table*}

Figure \ref{fig3} shows the similarity of each class to the other classes on Fashion MNIST and CIFAR-10, and is sorted in descending order of similarity. In Table \ref{table4}, we list in detail the name of each class in Figure \ref{fig3}a and Figure \ref{fig3}b.

\begin{figure*}[t]
\vskip -0.1in
\centering
\includegraphics[width=6in]{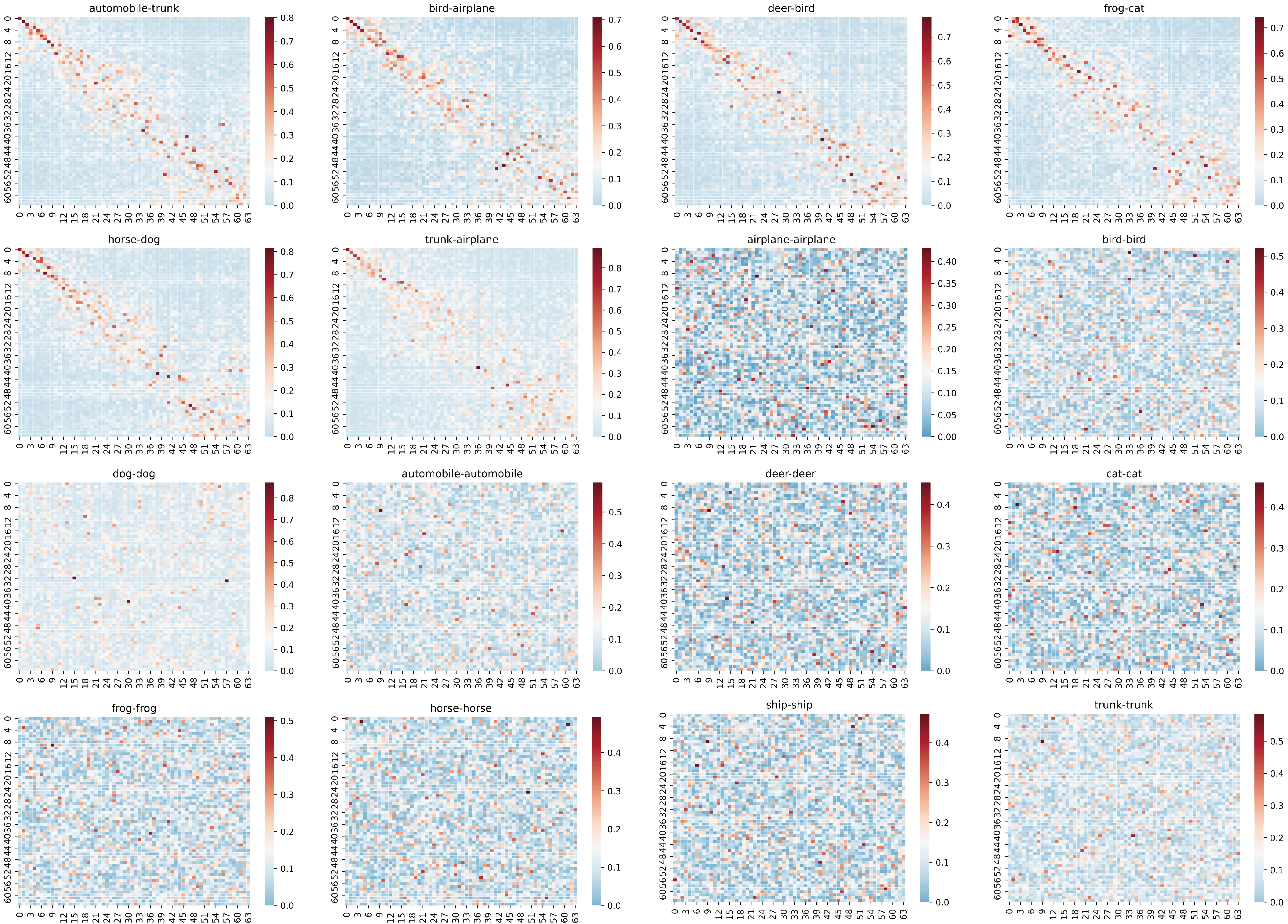}
\vskip 0.05in
\caption{Other examples with high geometric similarity in CIFAR-10 and geometry similarity between the same class of feature distributions extracted by different models.}
\label{fig11}
\end{figure*}

In addition to the geometry similarity between the feature distributions of \emph{dog} and \emph{cat} shown in Figure \ref{fig3}c, we also plot the geometry similarity between other classes with high similarity in Figure \ref{fig11}. This evidence strongly suggests that our findings are not accidental.

\section*{Appendix E}\label{secE}

In this section, we provide additional experimental results for phenomenon 3. Features of all classes in CIFAR-10 were extracted using two ResNet-18 trained with different initialization parameters, and then the geometry similarity between the feature distributions of the same class extracted by different models was calculated. All additional experimental results are plotted in Figure \ref{fig11}, where it can be observed that the same class of features extracted by the different models does not match phenomenon 2 at all.




\end{appendices}

\bibliographystyle{plain}
\bibliography{egbib}

\end{sloppypar}

\end{document}